\documentclass[]{jair}

\usepackage{tikz}
\usetikzlibrary{arrows}

\newtheorem{thm}{Theorem}[section]
\newtheorem{propo}[thm]{Proposition}
\newtheorem{lem}[thm]{Lemma}
\newtheorem{example}[thm]{Example}
\newtheorem{cor}[thm]{Corollary}
\theoremstyle{definition}
\newtheorem{defn}[thm]{Definition}

\newtheorem{rem}[thm]{Remark}

\newcommand{\Interpretations}{\mathcal{I}}
\newcommand{\Valuations}{\mathcal{V}}

\newcommand{\Lang}{\mathbb{L}}
\newcommand{\Logic}{\mathcal{L}}
\newcommand{\Pset}{\mathbb{P}}
\newcommand{\LogRepr}{\mathcal{R}}

\newcommand{\BeliefSets}{\mathcal{K}}
\newcommand{\ClosedSets}{\mathscr{C}}
\newcommand{\ClopenSets}{\mathscr{C}^\circ}

\newcommand{\MaxConsSets}{\mathcal{M}_\Logic}
\newcommand{\PVtoPL}{\mathcal{T}}
\newcommand{\PLtoPV}[1][\cdot]{\left\| #1 \right\|}
\newcommand{\evdot}{(\cdot)}

\newcommand{\opening}{\downarrow}
\newcommand{\remainder}{{\, \bot \,}}
\newcommand{\setMax}{\textit{Max}\,}

\newcommand{\generico}{\bullet}
\newcommand{\expn}{+}
\newcommand{\bigplus}{\mathop{\vcenter{\hbox{\huge $+$}}}\nolimits}\newcommand{\cont}{-}
\newcommand{\rev}{\ast}

\newcommand{\clr}{\mathbin{\scalebox{0.7}{$\copyright$}}}
\newcommand{\vers}{\circledcirc}

\newcommand{\rel}{\angle}
\newcommand{\Cred}{\prec}
\newcommand{\Quest}{\succ}
\newcommand{\FullQuest}{{\succ \! \succ}}

\newcommand{\si}{\mbox{If }}
\newcommand{\ssi}{\mbox{ iff }}
\newcommand{\cc}{\mbox{Otherwise}}
\newcommand{\tq}{\, | \,}
\newcommand{\tonce}{\mbox{ then }}

\newcounter{postcounter}
\newenvironment{postulate}[1]
    {\list{}
     {\setlength{\leftmargin}{2.5em}      % Indentation for whole item
      \setlength{\itemindent}{-2.5em}     % Negative indent for label
      \setlength{\labelwidth}{2em}        % Space for label
      \setlength{\labelsep}{0.5em}        % Space between label and text
      \setlength{\listparindent}{0pt}     % No paragraph indentation
      \setlength{\parsep}{0.5ex}          % Paragraph spacing
      \setlength{\itemsep}{0.5ex}         % Item spacing
      }        
     \renewcommand{\thepostcounter}{(#1\arabic{postcounter})}
     \let\olditem\item
     \renewcommand\item{\refstepcounter{postcounter}%
                       \olditem[]\textbf{\thepostcounter} \phantomsection\ignorespaces}}
    {\endlist
     \setcounter{postcounter}{0}}

\newcommand{\appendixcontent}{}

\makeatletter
\long\def\beginpraax#1#2#3\endpraax{%
    \g@addto@macro\appendixcontent{%
        \begingroup
        \begin{proof}[Proof of #1 \autoref{#2}]%
        \label{proof:#2}%
        #3
        \end{proof}
        \endgroup
    }%
    \textit{Proof in Appendix, page \hyperlink{page.\getpagerefnumber{proof:#2}}{\getpagerefnumber{proof:#2}}.} %
}
\makeatother

\setcopyright{cc}
\RequirePackage[
  datamodel=acmdatamodel,
  style=acmauthoryear,
  backend=biber,
  giveninits=true,
  uniquename=init
  ]{biblatex}

\begin{document}

%\title{JAIR Example Template}
\title[AWS Framework for Belief Change]
{An Abstract Worlds Semantic Framework for Belief Change Operators}
%% used to denote shared contribution to the research and/or corresponding author.

\author{Daniel Grimaldi}
\authornote{Corresponding Author.}
\authornote{All authors contributed equally.}
\orcid{0009-0000-6131-2372}
\email{dgrimaldi@dc.uba.ar}

\affiliation{%
  \institution{Departamento de Computación}
  \institution{Facultad de Ciencias Exactas y Naturales}
  \institution{Universidad de Buenos Aires}
  \city{Buenos Aires}
  \country{Argentina}
}

\affiliation{%
  \institution{Instituto de Investigación en Ciencias de la Computación}
  \institution{UBA-CONICET}
  \city{Buenos Aires}
  \country{Argentina}
}

\author{M. Vanina Martinez}
\orcid{0000-0003-2819-4735}
\email{vmartinez@iiia.csic.es}
\authornotemark[2]
\affiliation{%
  \institution{Artificial Intelligence Research Institute (IIIA-CSIC)}
  \city{Bellatera}
  \state{Cataluña}
  \country{Spain}}

\author{Ricardo O. Rodriguez}
\orcid{0000-0001-7551-2877}
\email{ricardo@dc.uba.ar}
\authornotemark[2]
\affiliation{%
  \institution{Departamento de Computación}
  \institution{Facultad de Ciencias Exactas y Naturales}
  \institution{Universidad de Buenos Aires}
  \city{Buenos Aires}
  \country{Argentina}
}

\affiliation{%
  \institution{Instituto de Investigación en Ciencias de la Computación}
  \institution{UBA-CONICET}
  \city{Buenos Aires}
  \country{Argentina}
}

%% The short list of authors must be made of the list of all authors' lastnames.
\renewcommand{\shortauthors}{Grimaldi, Martinez  \&  Rodriguez}

\begin{abstract}
This article proposes a set-theoretic framework for belief change, called Abstract Worlds Semantics, in which no logical syntax is assumed.
Inspired by Grove’s (1988) results, our approach treats worlds as primitive elements, over which world contraction and world revision operators are defined.
This semantic framework enables a unified analysis of belief change models.
Within this framework, we unify classical and non-prioritized belief change constructions by defining versatile operators.
When classical propositional logic is considered, our framework provides a homogeneous account of AGM, KM, and Multiple Change models.
In summary, AWS systematizes belief change frameworks and operators, simplifying and generalizing belief change theory over belief sets.
\end{abstract}
%% JAIR Note: 

%% To be updated by authors.
\received{20 February 2007}
\received[accepted]{5 June 2009}

\maketitle

\section{Introduction}
The seminal work of AGM~\cite{AGM1985}, a cornerstone of belief change theory, inherently proposes the idea that belief change operators should be definable for any logic~\cite{flouris2006}.
However, further development has led to a diverse set of proposals for operators and frameworks, most of them tailored to Classical Propositional Logic (CPL).
Most notable frameworks, aside from classical AGM, are Katsuno-Mendelzon~\cite{KM1990} and Multiple Change~\cite{Fuhrmann1988}.
The operators in these frameworks often remain tied to their underlying structure, making cross-framework comparison and adaptation difficult.
Meanwhile, studies on non-classical logics typically consider only classical AGM operators~\cite{flouris2006, Testa2017Paracons, RNW18nonCompact}, unable to account for the remaining variants.

To address this limitation, we introduce a set-theoretic framework called \textit{Abstract Worlds Semantics} (AWS), based on Grove’s possible world semantics~\cite{Grove1988}, where ``worlds'' are treated as primitive elements, independent of any specific logic.
This framework captures the core dynamics of the AGM theory in a purely semantic form, enabling the joint analysis of classical and non-prioritized operators from AGM, KM, and Multiple Change simultaneously.
By deliberately evading the association with a logic, we can develop a formal connection between the semantic structure of AGM and the theories of a Tarskian logic. 
This allows us to recover the classical frameworks in the CPL case while analyzing AGM-compliance~\cite{flouris2006} for non-prioritized operators and the diversity of classical operators in other Tarskian logics.
Within this setting, we also identify a new family of operators, called \textit{versatile}, unifying known operators under a single abstract construction.
In summary, AWS systematizes belief change frameworks and operators, simplifying and generalizing belief change theory over belief sets.

The paper is structured as follows.
Section \ref{sec:abstract-world-semantic} introduces the AWS framework and the versatile operators, providing axiomatic definitions, constructions, and representation theorems.
By recovering key concepts \cite{Lindstrom2022, TextbookHansson1999}, Section \ref{sec:AWS-applied-TLogics} shows how our proposal applies to Tarskian logics, and Section \ref{sec:mc-versatile-op} links AWS to AGM, KM, and Multiple Change frameworks.
Finally, in Section~\ref{sec:conclusions}, we summarize our results and future directions of research.
The ideas underlying this paper were first developed in Grimaldi's doctoral thesis~\cite{GrimTesis2025}, but the present paper expands the formal connection between the AWS framework and Tarskian logics, providing a more rigorous and general account of its semantics.

\section{The Abstract Worlds Semantics Framework}\label{sec:abstract-world-semantic}

The formal setting of the Abstract Worlds Semantic (AWS) framework is the algebra of an arbitrary index set denoted by $\Interpretations$, namely $(\Pset(\Interpretations), \cap, \cup, ^c, \subseteq)$\footnote{$X^c = \{ w \in \Interpretations \tq w \not \in X \}$ is the complement set of $X$ over $\Interpretations$.}.
Working with an algebra of sets gains in clarity and intuition, and can be interpreted as a generalization of Grove's possible world representation~\cite{Grove1988}.
In fact, the concepts developed in this work are inspired by it, for example, by calling \textit{worlds} to the elements of $\Interpretations$.

This specific form of Boolean algebra may be seen as a particular case of the AGM theory, since it resembles the finite case of CPL.
However, this simplicity is a feature rather than a limitation: by considering the concept of worlds as primitive, we avoid the association with a logic.
This key abstraction allows us in Section \ref{sec:AWS-applied-TLogics} to show that our framework is general enough to cover more than the arbitrary CPL case.
In this section, we show that the AWS framework can semantically capture the belief change theory.

One of the principles of the possible world context is that both the old (existing) knowledge and the new information are interpreted as sets.
Moreover, the epistemic attitudes can be represented in terms of the intersection and inclusion of such sets.
In AWS, we take this representation as primitive, i.e., given $A$, $B \subseteq \Interpretations$ where $A$ represents the knowledge and $B$ an information, we say that:
\begin{itemize}
    \item $B$ is believed if $A \subseteq B$;
    \item $B$ is disbelieved if $A \subseteq B^c$ and $A\neq \emptyset$;
    \item $B$ is unsettled if $A \cap B \neq \emptyset \neq A \cap B^c$.
\end{itemize}
As usual, a change operator takes the knowledge in the first parameter, the information in the second parameter, and the result represents the new knowledge.

\begin{figure}[!ht]
\centering
    \begin{tikzpicture}[node distance=1cm, align=center, auto, yscale=0.8,xscale=1]
        \node at (0,1.5) {$B$ unsettled};
        \draw (-1.8,-1) -- (-1.8,1) -- (1.8,1) -- (1.8,-1) -- (-1.8,-1);
        \draw [thick] (0.5,0) circle (0.75);
        \node at (0.75,0) {$B$};
        \draw [thick] (-0.5,0) circle (0.75);
        \node at (-0.75,0) {$A$};

        \node at (-4,-4.5) {$B$ is believed};
        \draw (-5.8,-4) -- (-5.8,-2) -- (-2.2,-2) -- (-2.2,-4) -- (-5.8,-4);
        \draw [thick] (-4,-3) circle (0.4);
        \node at (-4,-3) {$A$};
        \draw [thick] (-4,-3) circle (0.9);
        \node at (-4.65,-3) {$B$};

        \node at (4,-4.5) {$B$ is disbelieved};
        \draw (5.8,-4) -- (5.8,-2) -- (2.2,-2) -- (2.2,-4) -- (5.8,-4);
        \draw [thick] (3.1,-3) circle (0.75);
        \node at (3.1,-3) {$A$};
        \draw [thick] (4.9,-3) circle (0.75);
        \node at (4.9,-3) {$B$};
    
        \draw [-triangle 45] (2.2,-3) to [out=180, in=0] (-2.2,-3);
        \node at (0,-2.6) {$A \rev_w B$};
        \draw [-triangle 45] (-2.2,-3.8) to [out=0, in=180] (2.2,-3.8);
        \node at (0,-3.4) {$A \rev_w B^c$};

        \draw [-triangle 45] (-1.75,0.5) to [out=180, in=90] (-5,-2);
        \node at (-4,0.5) {$A \expn_w B$};
        \draw [-triangle 45] (1.75,0.5) to [out=0, in=90] (5,-2);
        \node at (4,0.5) {$A \expn_w B^c$};

        \draw [-triangle 45] (2.2,-2.4) to [out=180, in=270] (0.5,-1);
        \node at (1.9,-1.6) {$A \cont_w B^c$};
        \draw [-triangle 45] (-2.2,-2.4) to [out=0, in=270] (-0.5,-1);
        \node at (-1.9,-1.6) {$A \cont_w B$};
    \end{tikzpicture}
    \caption{Triangle of epistemic attitudes for AWS, based on the reference book \cite{FH2018}. We make a notation abuse by using $\expn_w$, $\cont_w$, and $\rev_w$ to represent \textbf{Expansion}, \textbf{Contraction}, and \textbf{Revision} movements respectively, since the associated operators are yet to be defined.
    The $w$ subindex emphasizes the AWS context.
    \Description{A diagram with three main boxes and arrows. Top box labeled ``B unsettled'' contains two overlapping circles: A on the left, B on the right. Bottom left box labeled ``B is believed'' contains a small circle A inside a larger circle B. Bottom right box labeled ``B is disbelieved'' contains two separate circles: A on the left, B on the right. Arrows: from top box to bottom left, the expansion of A and B; from top box to bottom right, the expansion of A and the complement of B. Between the two bottom boxes, an upper arrow from left to right with A revised by B, and a lower arrow from right to left labeled A revised by the complement of B. Also, arrows from top box directly to bottom left labeled A contracted by B, and to bottom right labeled A contracted by the complement of B^c.}
    }\label{fig:AWS-attitude-triangle}
\end{figure}

This is enough to recover the diagram that summarizes the behavior~\cite{FH2018}, in terms of the epistemic attitudes, of \textbf{Expansion}, \textbf{Contraction}, and \textbf{Revision} (see Figure \ref{fig:AWS-attitude-triangle}).
Here we are considering the concept of \textit{movement} rather than operator, where a movement is a specific type of change for a given pair of knowledge and information that can be made by a belief change operator.
In the original operators of AGM, it is implicit that the expansion operator ($\expn$) only make an \textbf{Expansion} movement; while a contraction operator ($\cont$) only make a \textbf{Contraction} movement, except when the information is a tautology; analogously, a revision operator ($\rev$)  make a \textbf{Revision} movement for every inconsistent pair, and an \textbf{Expansion} movement in the consistent case.
This last situation is often depicted as revision operators being extensions of the expansion operator.

But other operators, such as shielded contraction~\cite{FH1999ShC}, credibility-limited revision~\cite{H2001} (CL revision for short), and filtered revision~\cite{Bonanno2020,garapa_2022} exhibit a \textbf{No Change} movement when the knowledge needs to be preserved.
While promotion~\cite{SKM2018} and any other moderated revision~\cite{GMR2024} admit a \textit{Moderation} movement, resulting in a new knowledge that partially accepts the information, and doubts the rest.
We use the following example~\cite{GMR2024} to show the differences between these five movements, and how to represent them in the AWS framework.

\begin{example}\label{xej:clemenina-loteria-non-prioritized-revision}
After winning the lottery, Clementina's ex-partner calls, saying that s/he loves her and has always regretted breaking up.
However, about a month ago, s/he said that s/he did not love her anymore.
We represent the options Clementina has to change her beliefs with a finite CPL by considering the following propositional variables:

\begin{description}
    \item[$p$] ``The ex-partner has always regretted breaking up'' 
    \item[$q$] ``The ex-partner loves her'' 
\end{description}
Possible worlds here represent the complete formulas of the logic as tuples:
\[
\begin{array}{lcl}
    (1,1) \mbox{ represents } p \wedge q   & \qquad & (1,0) \mbox{ represents } p \wedge \neg q  \\
    (0,1) \mbox{ represents } \neg p \wedge q & \qquad &(0,0) \mbox{ represents } \neg p \wedge \neg q 
\end{array}
\]

Therefore, $\Interpretations = \{ (1,1), (1,0), (0,1), (0,0) \}$.
Clementina's original knowledge is $\psi = \neg q$, associated to the set $\{ (1,0) , (0,0)\}$.
The information $\mu= p \wedge q$ is what she received by phone, whose set is $\{ (1,1) \}$.
We now analyze her alternatives:
\begin{description}
    \item[No Change] Clementina preserves her knowledge, acting as if the call never happened. Therefore, her possible worlds are still $\{ (1,0), (0,0)\}$.
    \item[Expansion] Clementina decides to adopt the new information without ensuring consistency.
    She now simultaneously believes in $q$ and $\neg q$, and no world can represent this.
    Her new knowledge is represented by the empty set.
    \item[Revision] Clementina decides to adopt the new information consistently.
    But since $\mu$ is a complete formula, her new knowledge is the only non-empty subset of $\{ (1,1) \}$, i.e. $\{ (1,1) \}$ itself.
    She now believes the new information.
    \item[Contraction] Clementina decides not contradict the information\footnote{Here \textbf{Contraction} is seen as taking the negation of the new information~\cite{SKM2018, GMR2024}.}.
    She preserves all her worlds and also adds the one from the information.
    Thus, $\{ (1,1), (1,0), (0,0)\}$ represents her new knowledge.
    Now she only believes that if her ex-partner loves her, then s/he has always regretted breaking up.
    \item[Moderation] Clementina accepts changing her beliefs according to the information, but she does it by adopting some affirmations and doubting others.
    A possible set of worlds is $\{ (1,1), (1,0) \}$, where she now believes her ex-partner has always regretted breaking up, but doubts whether s/he loves her.
    Another alternative is $\{ (1,1), (0,0)\}$, doubting about the true feelings of her ex-partner, but understanding his/her regret as equivalent to loving her.
\end{description}
\end{example}

According to the example, in the AWS framework, we represent these movements by selecting elements from two sets: $A$, the knowledge, and $B$, the new information.
In each case, the new knowledge $C$ is interpreted differently:
\begin{description}
    \item[No Change] $C$ is just $A$;
    \item[Expansion] $C$ is $A \cap B$;
    \item[Revision] $C$ is the union of $A \cap B$ and some worlds from $A^c \cap B$;
    \item[Contraction] $C$ is the union of $A$ and some worlds from $A^c \cap B^c$;
    \item[Moderation] $C$ has some worlds from $A^c \cap B$ and $A \cap B^c$, and all from $A \cap B$.
\end{description}
As stated before, some operators can combine these movements, depending on the pair of knowledge and the information given.
Thus, these interpretations extend the notion of change to cover the behavior of non-prioritized operators.
The aim now is to develop a change operator defined in AWS where all these movements shape the possible outcomes for a given $(A, B)$.
In the following subsections, we present the set-theoretic formalization with an AGM methodology: postulates, constructions, and representation theorems.

\subsection{Elementary World Operators}%%%%%%%%%%%%%%%%%%%%%%%%%%%%%%%%%%%%%%%%%%%%%%

When defining operators, both constructively and by postulates, we based them on the underlying algebra of sets.
This observation is crucial when, in Section \ref{sec:AWS-applied-TLogics}, we apply our framework to a given Tarskian logic.
Therefore, we avoid using the unary complement operator $^c$ in the postulates, so they can be easily translated in contexts where it may not be suitably defined, such as Multiple Change.

Following the spirit of adapting from the possible world semantics, constructing operators is based on world-selection functions~\cite{Grove1988, TextbookHansson1999}.
In the AWS framework, we consider both knowledge and information as parameters for the new knowledge.
Thus, our world-selection function has two parameters.

\begin{defn} \label{def:world-selection-function}
The function $\sigma: \Pset(\Interpretations) \times \Pset(\Interpretations) \longrightarrow \Pset(\Interpretations)$ is defined as a \textit{world-selection} if $\sigma(A, B) \subseteq (A \cup B)^c$.
\end{defn}

World-selection functions identify which new worlds are being considered for the new knowledge.
With this general definition, we can already formally define the \textit{elementary} versions of world contraction and revision operators.

\begin{defn}\label{def:world-contraction-revision-op}
    We call \textit{world} operators, using the $w$ subindex, as the ones defined over $\Pset(\Interpretations)$, i.e. $\generico_w: \Pset(\Interpretations) \times \Pset(\Interpretations) \longrightarrow \Pset(\Interpretations)$.
    The world operators $\cont_w$, $\rev_w$ are called \textit{elementary world contraction} and \textit{elementary world revision} respectively, if there is a world-selection function $\sigma$ such that given $A$, $B \in \Pset(\Interpretations)$:
    \[
    A \cont_w B = A \cup \sigma(A,B) \qquad \qquad A \rev_w B = (A \cap B) \cup \sigma(A,B^c)
    \]
    We may write $\cont_w^\sigma$ and $\rev_w^\sigma$ to refer to the operators that are constructed by $\sigma$.
    When an elementary world contraction and revision are constructed by the same world-selection function, we say that one operator is the \textit{dual} of the other.
\end{defn}

An interesting particular case happens when the world-selection function is constantly the empty set ($\sigma(A, B)=\emptyset$).
The associated elementary world contraction only has the \textbf{No Change} movement, i.e., it only returns $A$: the world constant operator.
Its dual elementary world revision only has the \textbf{Expansion} movement; therefore, it is the world expansion operator.
In AWS, the world expansion operator is also identified with the intersection of the algebra of sets.
In general, elementary world contraction operators cover \textbf{No Change} and \textbf{Contraction} movements, while
elementary revision operators cover \textbf{Expansion} and \textbf{Revision} movements.
Such operators are definable in terms of postulates, and they also satisfy Levi and Harper Identities.

\begin{thm}\label{thm:world-contraction-revision-repr-and-identitities}
Let $\cont_w$, $\rev_w$ be two world operators.
\begin{enumerate}
    \item Operator $\cont_w$ is an elementary world contraction iff it satisfies:
\begin{postulate}{W$\cont$}
    \item $A \subseteq A \cont_w B$ (world contraction inclusion). \label{post:world-c-inclusion}
    \item $(A \cont_w B) \cap B \subseteq A$ (world recovery). \label{post:world-c-recovery}
\end{postulate}
    \item Operator $\rev_w$ is an elementary world revision iff it satisfies:
\begin{postulate}{W$\rev$}
    \item $A \rev_w B \subseteq B $ (world revision success). \label{post:world-r-success}
    \item $A \cap B \subseteq A \rev_w B$ (world inclusion). \label{post:world-r-inclusion}
\end{postulate}
    \item If $\cont_w$, $\rev_w$ are elementary world contraction and revision operators constructed by the same world-selection function $\sigma$, then they satisfy the Levi and the Harper Identities:
\begin{description}
    \item[Levi Identity] $A \rev_w^\sigma B = (A -_w^\sigma B^c) \cap B$
    \item[Harper Identity] $A -_w^\sigma B = (A \rev_w^\sigma B^c) \cup A$
\end{description}
\end{enumerate}
\end{thm}

\beginpraax{Theorem}{thm:world-contraction-revision-repr-and-identitities}
\begin{enumerate}
    \item[] 
    \item Consider an operator $\cont_w$ satisfying \ref{post:world-c-inclusion} and \ref{post:world-c-recovery} and define the function $\sigma(A,B)=(A \cont_w B) \cap A^c$.
    Then, $\sigma(A,B) \cap B = (A \cont_w B) \cap A^c \cap B$.
    By \ref{post:world-c-recovery} we have $\sigma(A,B) \cap B \subseteq A \cap A^c = \emptyset$.
    Hence, $\sigma(A,B) \cap B = \emptyset$, which is equivalent to $\sigma(A,B) \subseteq B^c$.
    Therefore $\sigma(A,B) \subseteq A^c \cap B^c$, i.e., $\sigma$ is a world-selection function.

    Given now a world-selection function $\sigma$, consider the operator defined as $A \cont_w B = A \cup \sigma(A, B)$.
    Then $\cont_w$ satisfies \ref{post:world-c-inclusion} by definition and since $\sigma(A , B) \subseteq B^c$, we have $(A \cont_w B) \cap B = A \cap B$.
    Thus, $\cont_w$ satisfies \ref{post:world-c-recovery}.
    
    \item Consider an operator $\rev_w$ satisfying \ref{post:world-r-success} and \ref{post:world-r-inclusion} and define the function $\sigma(A,B)=(A \rev_w B^c) \cap A^c$.
    By \ref{post:world-r-success} we have $\sigma(A,B) \subseteq B^c \cap A^c$, therefore $\sigma$ is a world-selection function.

    Given now a world-selection function $\sigma$, consider the operator defined as $A \rev_w B = (A \cap B) \cup \sigma(A, B^c)$.
    Then $\rev_w$ satisfies \ref{post:world-r-success} since $\sigma(A , B^c) \subseteq B$ and it satisfies \ref{post:world-r-inclusion} by construction.

\item Let us check the identities using the constructions:

\noindent
\begin{tabular}{@{}l@{}rl@{}}
\textbf{Levi Identity: } & $A \rev_w B$ & $= (A \rev_w B) \cap B = (A \cap B) \cup (\sigma(A, B^c) \cap B)$ \\
    & &$= (A \cup \sigma(A, B^c)) \cap B = (A -_w B^c) \cap B$ \\
\textbf{Harper Identity: } & $A -_w B $ & $= A \cup \sigma(A, B) = A \cup (A \cap B^c) \cup \sigma(A, B)$ \\
& & $= A \cup (A \rev_w B^c)$ \\
\end{tabular}

\end{enumerate}
\endpraax %% END OF PROOF

Elementary operators are more general than classical ones.
On one side, there is no vacuity situation considered.
On the other side, there is a chance of failing the contraction, and revision may be inconsistent.
The inclusion postulates \ref{post:world-c-inclusion} and \ref{post:world-r-inclusion} characterize the worlds of the original knowledge that has to be preserved for the new knowledge.
While postulates revision success \ref{post:world-r-success} and recovery \ref{post:world-c-recovery}, by considering the equivalent statement $A \cont_w B \subseteq A \cup B^c$, restrict the new worlds to be added.
Through these operators, we cover all the movements but \textbf{Moderation}.

Following the definition of moderated revision~\cite{GMR2024}, \textbf{Moderation} movement represents what ``lies between'' revision and contraction, where contraction is understood as applied to the negation of the information.
By looking for an operator not only able to make \textbf{Moderation} movement, but also able to behave like elementary world contractions and revisions, we are challenging a core principle in the belief change theory: the duality between contraction and revision operators, given by the Levi and Harper Identities.

From postulates \ref{post:world-c-inclusion} and \ref{post:world-c-recovery}, we have that an elementary contraction operator $\cont_w$ satisfies $ A \subseteq A \cont_w B^c \subseteq A \cup B$.
This defines a range of possibilities that lies between $A$ and $A \cup B$.
Analogously, postulates \ref{post:world-r-success} and \ref{post:world-r-inclusion} affirms that $ A\cap B \subseteq A \rev_w B \subseteq B $ for any elementary revision operator $\rev_w$.
Thus, the operators that ``lie between'' revision and contraction should return sets from these ranges.
We call these operators \textit{versatile}.

\begin{defn}\label{def:world-versatile}
    A world operator $\vers_w$ is an elementary world versatile if, given $A, B \in \Pset(\Interpretations)$, it satisfies \ref{post:world-r-inclusion} and:
\begin{postulate}{W$\vers$}
    \item \label{post:world-shared-success} $A \vers_w B \subseteq A \cup B$ (shared success).
\end{postulate}
\end{defn}

Thus, the range of such an is $ A\cap B \subseteq A \vers_w B \subseteq A \cup B$.
But this implies more than covering the ranges of the other two elementary world operators.

\begin{lem}\label{lem:world-shared-success-vs-harper-identity}
    Let $\vers_w$ be a world operator. The following are equivalent:
    \begin{enumerate}
        \item $\vers_w$ satisfies \ref{post:world-shared-success};
        \item $\cont_w$, defined as $A \cont_w B = A \cup (A \vers_w B^c)$, is an elementary world contraction;
        \item $\cont_w'$, defined as $A \cont_w' B = A \cup (B^c \vers_w A)$, is an elementary world contraction.
    \end{enumerate}
    We name the elementary world contractions $\cont_w$ and $\cont_w'$ as associated with $\vers_w$.
\end{lem}

\beginpraax{Lemma}{lem:world-shared-success-vs-harper-identity}
\textbf{From $\vers_w$ to $\cont_w$ and $\cont_w'$:}
    Operators $\cont_w$ and $\cont_w'$ satisfy \ref{post:world-c-inclusion} by construction.
    Meanwhile, $A \cont_w B$ and $A \cont_w' B$ are included in $A \cup B^c$ since $\vers_w$ satisfies \ref{post:world-shared-success}.
    Hence, intersecting by $B$ implies $\cont_w$ and $\cont_w'$ satisfy \ref{post:world-c-recovery}.

\textbf{From $\cont_w$ or $\cont_w'$ to $\vers_w$:} By rewriting the construction of $\cont_w$ and $\cont_w'$ both in terms of $A \vers_w B$, we have that:
    \[
    A \cont_w B^c = A \cup (A \vers_w B)  \qquad B \cont_w' A^c = B \cup (A \vers_w B)
    \]
    By the recovery postulate \ref{post:world-c-recovery}, satisfied by $\cont_w$ and $\cont_w'$, we deduced that:
    \begin{gather*}
        (A \cont_w B^c) \cap B^c =  (A \cup (A \vers_w B)) \cap B^c \subseteq A  \\
        (B \cont_w' A^c) \cap A^c = (B \cup (A \vers_w B))\cap A^c \subseteq B
    \end{gather*}
    Thus if $\cont_w$ holds \ref{post:world-c-recovery}, then $(A \vers_w B) \cap B^c \subseteq A$; and if $\cont_w'$ holds \ref{post:world-c-recovery}, then $(A \vers_w B) \cap A^c \subseteq B$.
    In any case, $A \vers_w B \subseteq A \cup B$, i.e. \ref{post:world-shared-success} holds.
\endpraax

Hence, all the operators characterized by postulate \ref{post:world-shared-success} are associated, via Harper, to elementary world contraction operators. 
The same occurs with elementary world revision operators, Levi construction, and postulate \ref{post:world-r-inclusion}. 

\begin{lem}\label{lem:world-r-inclusion-vs-levi-identity}
    Let $\vers_w$ be a world operator. The following are equivalent:
    \begin{enumerate}
        \item $\vers_w$ satisfies \ref{post:world-r-inclusion};
        \item $\rev_w$, defined as $A \rev_w B = (A \vers_w B) \cap B$, is an elementary world revision;
        \item $\rev_w'$, defined as $A \rev_w' B = (B \vers_w A) \cap A$, is an elementary world revision.
    \end{enumerate}
    We name the elementary world revisions $\rev_w$ and $\rev_w'$ as associated with $\vers_w$.
\end{lem}

\beginpraax{Lemma}{lem:world-r-inclusion-vs-levi-identity}
\textbf{From $\vers_w$ to $\rev_w$ and $\rev_w'$:}
    Operators $\cont_w$ and $\cont_w'$ satisfy \ref{post:world-r-success} by construction.
    Meanwhile, since $\vers_w$ satisfies \ref{post:world-r-inclusion}, both $A \rev_w B$ and $A \rev_w' B$ includes $A \cap B$.
    Hence, $\rev_w$ and $\rev_w'$ satisfy \ref{post:world-r-inclusion}.

\textbf{From $\cont_w$ or $\cont_w'$ to $\vers_w$:} By rewriting the construction of $\rev_w$ and $\rev_w'$ both in terms of $A \vers_w B$, we have that:
\[
        A \rev_w B = (A \vers_w B) \cap B \subseteq A \vers_w B \qquad
        B \rev_w' A = (A \vers_w B) \cap A \subseteq A \vers_w B
\]
    Thus, if $\rev_w$ or $\rev_w'$ satisfies \ref{post:world-r-inclusion}, then $\vers_w$ also satisfies it.
\endpraax

When combined with the representation theorems of elementary world contraction and revision, these lemmas lead us to the representation theorem of elementary world versatile operators.

\begin{thm}\label{thm:world-versatile-separation-representation}
    Let $\vers_w$ be a world operator. The following are equivalent:
    \begin{enumerate}
        \item The operator $\vers_w$ satisfies \ref{post:world-shared-success} and \ref{post:world-r-inclusion}.
        \item There are elementary world contractions $\cont_w$, $\cont_w'$ and their dual elementary world revisions $\rev_w$, $\rev_w'$, all associated to $\vers_w$, such that for every $A$, $B \subseteq \Interpretations$:
        \[
        A \vers_w B = (A \cont_w B^c) \cap (B \cont_w' A^c) = (A \rev_w B) \cup (B \rev_w' A)
        \]
        \item There are world-selection functions $\sigma$ and $\rho$ such that for every $A$, $B \subseteq \Interpretations$:
        \[
        A \vers_w B = (A \cap B) \cup \sigma(A, B^c) \cup \rho(B,A^c)
        \]
    \end{enumerate}
    In this case, we say that operator $\vers_w$ is constructed by $(\sigma, \rho)$, where $\rev_w$ and $\cont_w$ are constructed by $\sigma$ and $\rev_w'$ and $\cont_w'$ are constructed by $\rho$.
\end{thm}

\beginpraax{Theorem}{thm:world-versatile-separation-representation}
    From Lemma \ref{lem:world-shared-success-vs-harper-identity}, if $\vers_w$ also satisfies \ref{post:world-r-inclusion} we have:
    \[
    (A \cont_w B^c) \cap (B \cont_w' A^c) = (A \cap B) \cup (A \vers_w B) = A \vers_w B 
    \]
    Analogously, from Lemma \ref{lem:world-r-inclusion-vs-levi-identity}, if $\vers_w$ also satisfies \ref{post:world-shared-success} we have:
    \[
    (A \rev_w B) \cup (B \rev_w' A) = (A \cup B) \cap (A \vers_w B) = A \vers_w B 
    \]

    Assuming these constructions, we can apply the representation Theorem \ref{thm:world-contraction-revision-repr-and-identitities} that allows us to represent the $\vers_w$ operator using world-selection functions.
    Consider $\rev_w$ and $\rev_w'$ constructed by $\sigma$ and $\rho$, then:
    \[
    A \vers_w B = (A \cap B) \cup \sigma(A,B^c) \cup \rho(B,A^c) 
    \]
    Given this representation, note that $A \cont_w B = A \cup (A \vers_w B^c) = A \cup \sigma(A,B)$ since $(A \cap B^c) \cup \rho(B^c,A^c) \subseteq A$.
    Analogously, $A \cont_w' B = A \cup (B^c \vers_w A) = A \cup \rho(A,B)$ since $(A \cap B^c) \cup \sigma(B^c,A^c) \subseteq A$.
    Hence, the elementary world contractions associated with $\vers_w$ are duals of the elementary world revisions associated with $\vers_w$.

    Lastly, it is directly deduced from the constructive representation of $\vers_w$ using world-selection functions that the operator satisfies \ref{post:world-shared-success} and \ref{post:world-r-inclusion}.
\endpraax

For elementary world versatile operators, the associated elementary world contractions and revisions can be seen as the extreme positions w.r.t. the inclusion, a property originally related to moderated revision operators:
\[
    A \rev_w^\sigma B \subseteq A \vers_w B \subseteq A \cont_w^\sigma B^c \qquad
    A \rev_w^\rho B \subseteq B \vers_w A \subseteq A \cont_w^\rho B^c
\]
But the flexibility of elementary versatile operators also allows us to see elementary world contraction and revision operators as subfamilies of versatile operators.

\begin{cor}\label{cor:rev-cont-as-vers}
Let $\vers_w$ be an elementary world versatile constructed by the pair $(\sigma, \rho)$, and $\cont_w^\sigma$ and $\rev_w^\sigma$ be its associated elementary contraction and revision.
Then, each line of the table represents equivalent statements for every $A$, $B \subseteq \Interpretations$:
\begin{center}
\begin{tabular}{@{}ccc@{}}
   \toprule
   \textbf{Postulate} & \textbf{world-selection functions} & \textbf{associated operators}\\
    \midrule
    \ref{post:world-c-inclusion} & $\rho(A,B)=A^c \cap B^c$ & $A \vers_w B^c = A \cont_w^\sigma B$\\ \addlinespace
    \ref{post:world-r-success} & $\rho(A,B)=\emptyset$ & $A \vers_w B = A \rev_w^\sigma B$\\
    \bottomrule 
\end{tabular}
\end{center}
\end{cor}

\beginpraax{Corollary}{cor:rev-cont-as-vers}
If \ref{post:world-c-inclusion} holds, then $A \cap (A \cap B^c) \subseteq (A \vers_w B) \cap (A \cap B^c)$.
Hence, by Theorem \ref{thm:world-versatile-separation-representation} and world-selection function definition, $A \cap B^c = \rho(B, A^c)$. Renaming the variables, we have $\rho(A,B)=A^c \cap B^c$.
Also, by Theorems \ref{thm:world-contraction-revision-repr-and-identitities} and \ref{thm:world-versatile-separation-representation}, if $\rho(A,B)=A^c \cap B^c$, then $A \vers_w B^c = A \cup \sigma(A,B) = A \cont_w^\sigma B$.
Lastly, if $A \vers_w B^c = A \cont_w^\sigma B$, then $\vers_w$ satisfies  \ref{post:world-c-inclusion}, since $\cont_w^\sigma$ also satisfies it.

If \ref{post:world-r-success} holds, then $(A \vers_w B) \cap (A \cap B^c) \subseteq B \cap (A \cap B^c)$.
Hence, by Theorem \ref{thm:world-versatile-separation-representation}, we deduce $\emptyset = \rho(B, A^c)$.
Renaming the variables, we have $\rho(A, B)=\emptyset$.
Also, by Theorem \ref{thm:world-contraction-revision-repr-and-identitities} and \ref{thm:world-versatile-separation-representation}, if $\rho(A,B)=\emptyset$, then $A \vers_w B = (A \cap B) \cup \sigma(A,B^c) = A \rev_w^\sigma B$.
Lastly, if $A \vers_w B = A \rev_w^\sigma B$, then $\vers_w$ satisfies \ref{post:world-r-success}, since $\rev_w^\sigma$ also satisfies it.
\endpraax

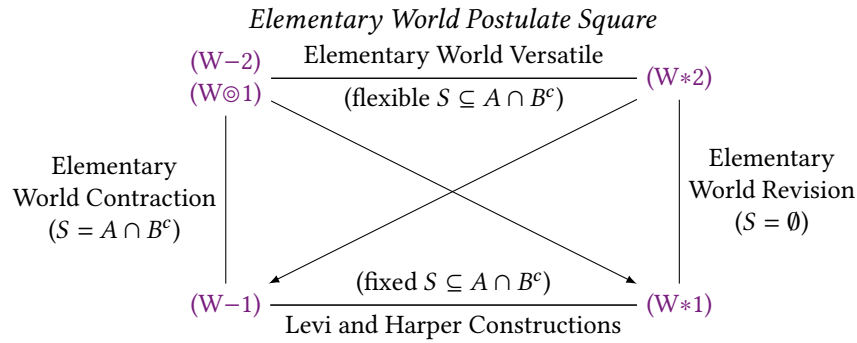
\begin{figure}[hbtp]
\centering
\begin{tikzpicture}[yscale=1.5,xscale=3, align=center, auto]
\node at (1,2.5) {\textit{\large Elementary World Postulate Square}};
  \node (ki) at (0,0) {\ref{post:world-c-inclusion}};
  \node (rs) at (2,0) {\ref{post:world-r-success}};
  \node (rec) at (0,2) {\ref{post:world-c-recovery}\\\ref{post:world-shared-success}};
  \node (ei) at (2,2) {\ref{post:world-r-inclusion}};
  \draw[-] (ki) to node [below] {Levi and Harper Constructions} (rs);
  \draw[-] (ki) to node [above] {(fixed $S \subseteq A \cap B^c$)} (rs);
  \draw[-] (rec) to node [above] {Elementary World Versatile} (ei);
  \draw[-] (rec) to node [below] {(flexible $S \subseteq A \cap B^c$)} (ei);
  \draw[-] (ki) to node [left] {Elementary\\World Contraction\\($S = A \cap B^c$)} (rec);
  \draw[-] (rs) to node [right] {Elementary\\World Revision\\($S = \emptyset$)} (ei);
  \draw[-latex] (rec) to node [rotate=45, above] {} (rs);
  \draw[-latex] (ei) to node [rotate=-45, below] {} (ki);
\end{tikzpicture}
\caption{This square of elementary postulates shows a spatial view of the relation between the elementary world operators and the Levi and Harper constructions.}\label{fig:elementary-postulates-square}
\Description{A directed and labeled graph where the nodes are the postulates: world contraction inclusion in the bottom left; world revision success in the bottom right; world contraction recovery and world shared success in the top left; world revision inclusion in the top right. The bottom edge is labeled ``Levi and Harper Constructions (fixed subset S of A intersect B complement)''. The top edge is labeled ``Elementary World Versatile (flexible subset S of A intersect B complement)''. The left edge is labeled ``Elementary World Contraction (S = A intersect B complement)''. The right edge is labeled ``Elementary World Revision (S = the empty set)''. Two diagonal arrows go from top left to bottom right and from top right to bottom left.}
\end{figure}

These results add a new dimension to the contraction-revision duality and the implication of the Levi and Harper constructions, as seen in Figure \ref{fig:elementary-postulates-square}.
When applying such constructions, the resulting operators can be seen as extreme positions with respect to elementary world versatile operators.
This is also seen in the postulates: \ref{post:world-r-success} implies \ref{post:world-shared-success}, and \ref{post:world-c-inclusion} implies \ref{post:world-r-inclusion}.
Thus, an elementary world revision operator can be seen as an elementary world versatile operator that avoids considering worlds from $A \cap B^c$, forced by \ref{post:world-r-success}.
Analogously, an elementary world contraction operator (evaluated in $B^c$) can be seen as an elementary world versatile operator that contains every world from $A \cap B^c$, forced by \ref{post:world-c-inclusion}.
Recall that an intermediate position, associated with a \textbf{Moderation} movement, is about considering some worlds from $A \cap B^c$.

In the following subsections, we define classical and non-prioritized operators in the AWS approach and how their associated world-selection functions behave.

\subsection{Classic World Operators}

Classical contraction and revision present extra properties than the ones satisfied by the elementary versions, namely, success for contraction, consistency for revision, and vacuity.

\begin{defn}\label{def:vacuity-supplementary-postulates}
     Given $A, B, B' \in \Pset(\Interpretations)$, consider the following postulates for elementary world contraction $\cont_w$:
    \begin{postulate}{W$\cont$}
        \setcounter{postcounter}{2}
        \item \label{post:world-c-success} If $B \neq \Interpretations$ then $A \cont_w B \not \subseteq B$ (world contraction success).
        \item \label{post:world-c-vacuity} If $A \not \subseteq B$ then $A \cont_w B = A$ (world contraction vacuity).
    \end{postulate}
    We call \textit{basic} world contraction when $\cont_w$ satisfies \ref{post:world-c-success} and \ref{post:world-c-vacuity}.
    
    Consider now the following postulates for elementary world revision $\rev_w$:
    \begin{postulate}{W$\rev$}
        \setcounter{postcounter}{2}
        \item \label{post:world-r-consistency} If $B \neq \emptyset$ then $A \rev_w B \neq \emptyset$ (world revision consistency).
        \item \label{post:world-r-vacuity} If $A \cap B \neq \emptyset$ then $A \rev_w B = A \cap B$ (world revision vacuity).
    \end{postulate}
    We call \textit{basic} world revision when $\rev_w$ satisfies \ref{post:world-r-consistency} and \ref{post:world-r-vacuity}.
\end{defn}

These postulates can be represented in terms of the world-selection function used to construct them.

\begin{thm}\label{thm:world-cont-rev-classic-repr}
Let $\rev_w$ be an elementary world revision constructed by $\sigma$ and $\cont_w$ its dual elementary world contraction.
Then, each line of the table represents equivalent statements for every set $A$, $B \subseteq \Interpretations$:
\begin{center}
\begin{tabular}{ccc}
   \toprule
   \textbf{$\cont_w$ postulate} & \textbf{$\rev_w$ postulate} & \textbf{$\sigma$ property} \\ \midrule
    \ref{post:world-c-success} &
    \ref{post:world-r-consistency} &
    If $A \subseteq B \subsetneq \Interpretations$ then $\sigma(A,B) \neq \emptyset$ \\
    & & (\textit{\textbf{semantically successful}})
    \\ \addlinespace
    \ref{post:world-c-vacuity} &
    \ref{post:world-r-vacuity} &
    If $A \not \subseteq B$ then $\sigma(A,B) = \emptyset$  (\textit{\textbf{normal}})
   \\\bottomrule
\end{tabular}
\end{center}
\end{thm}

\beginpraax{Theorem}{thm:world-cont-rev-classic-repr}
Let us check that a given property of $\sigma$ holds is equivalent to a postulate of the elementary world operators constructed by $\sigma$.

\textbf{$\sigma$ is semantically successful:} Suppose $\sigma$ is semantically successful and take $B \neq \Interpretations$.
    If $A \not \subseteq B$ then $A \cont_w^\sigma B \not \subseteq B$ by construction, and if $A \subseteq B$ then $\sigma(A, B) \neq \emptyset$ by hypothesis. 
    In any case, $A \cont_w^\sigma B \not \subseteq B$, hence $\cont_w^\sigma$ satisfies \ref{post:world-c-success}.
    If $\cont_w^\sigma$ satisfies \ref{post:world-c-success}, then $B \neq \Interpretations$ implies $(A \cup \sigma(A,B)) \cap B^c \neq \emptyset $ by construction.
    If $A \subseteq B \subsetneq \Interpretations$ we have that $\sigma(A,B)) \cap B^c \neq \emptyset$.
    Since $\sigma(A,B)) \subseteq B^c$ we conclude $\sigma(A,B)) \neq \emptyset$.
    Thus $\sigma$ is semantically successful.

    Suppose $\sigma$ is semantically successful, and take $B \neq \emptyset$.
    If $A \cap B \neq \emptyset$, then $A \rev_w B \neq \emptyset$ by construction, and if $A \cap B = \emptyset$ then $\sigma(A,B^c) \neq \emptyset$ by hypothesis.
    In any case, $A \rev_w B \neq \emptyset$, hence $\cont_w^\sigma$ satisfies \ref{post:world-r-consistency}.
    Assume now $\rev_w^\sigma$ satisfies \ref{post:world-r-consistency}, and take $A \subseteq B \subsetneq \Interpretations$.
    Then, since $B^c \neq \emptyset$, we have $(A \cap B^c) \cup \sigma(A,B) \neq \emptyset $ by construction.
    But $A \cap B^c = \emptyset$ by hypothesis, thus $\sigma(A,B) \neq \emptyset$.
    Therefore, $\sigma$ is semantically successful.

\textbf{$\sigma$ is normal:} Assume $\sigma$ is normal.
    If $A \not \subseteq B$, then $\sigma(A,B) = \emptyset$.
    By construction, $A \cont_w^\sigma B = A$, satisfying \ref{post:world-c-vacuity}.
    Analogously, if $A \not \subseteq B^c$, i.e. $A \cap B \neq \emptyset$, then $\sigma(A,B^c) = \emptyset$.
    Hence, $A \rev_w^\sigma B = A \cap B$, satisfying \ref{post:world-r-vacuity}.

    If $\cont_w^\sigma$ satisfies \ref{post:world-c-vacuity}, we deduce that $A \not \subseteq B$ implies $\sigma(A,B^c) = \emptyset$ by construction of $\cont_w^\sigma$, i.e., $\sigma$ is normal.
    Analogously, if $\rev_w^\sigma$ satisfies \ref{post:world-r-vacuity}, we deduce that $A \cap B \neq \emptyset$ implies $\sigma(A,B^c) = \emptyset$ by construction of $\rev_w^\sigma$.
    So, if $A \not \subseteq B'$, consider $A \cap {B'}^c \neq \emptyset$, thus $\sigma(A,B') = \emptyset$.
    Hence $\sigma$ is normal.
\endpraax

In the next subsection, non-prioritized concepts such as shielded, credibility, and moderation are depicted as properties of versatile operators.

\subsection{Credible and Questionable Relations}

In the AWS framework, we can analyze the elementary world versatile operators by separating $\Interpretations$ into four parts, given a knowledge $A$ and a new information $B$:
\begin{description}
    \item[$A\cap B$] has the worlds considered for both the knowledge and the new information: elementary world versatile always take them for the new knowledge;
    \item[$A^c\cap B$] has the worlds that ponder the new information over the original knowledge, giving its credibility: elementary world versatile represent this selection using the world-selection function $\sigma$;
    \item[$A \cap B^c$] has the worlds of the original knowledge that challenge the new information, thus partially questioning it: elementary world versatile represent this selection using the world-selection function $\rho$;
    \item[$A^c\cap B^c$] has the worlds that were never originally considered, neither the original knowledge nor the new information did it in the first time: elementary world versatile never consider them for the new knowledge;
\end{description}

Thus, given an elementary world versatile $\vers_w$ and a consistent knowledge $A$ ($A \neq \emptyset$), looking at $(A \vers_w B) \cap B = (A\cap B) \cup \sigma(A, B^c)$ is a way to understand the credibility towards an information $B$.
When the set is non-empty, it means that $B$ is credible.
Analogously, the questionability towards $B$ is represented by $(A \vers_w B) \cap B^c = \rho(B,A^c)$, where $\rho(B,A^c) \neq \emptyset$ means that $B$ is questionable.
We formalize and generalize this by defining credible and questionable relations associated with world operators.

\begin{defn}\label{def:world-change-perspective-relations}
    Let $\generico_w$ be a world operator over $\Pset(\Interpretations)$.
    We define the following relations associated with $\generico_w$:
    \begin{itemize}
        \item The relation $A \Cred_w B$ given by $(A \generico_w B) \cap B \neq \emptyset$ or $A=\emptyset$;
        \item The relation $A \Quest_w B$ given by $(A \generico_w B) \cap A \not \subseteq B$;
        \item The relation $A \FullQuest_w B$ given by $\emptyset \neq A \cap B^c \subseteq A \generico_w B$.
    \end{itemize}
\end{defn}

We identify the case $\emptyset \neq A \cap B^c \subseteq A \generico_w B$, because, if $\generico_w$ satisfies \ref{post:world-r-inclusion}, this means not only that $B$ is questionable, but also that the agent aims to preserve her/his knowledge or, at the very least, not to challenge the new information.

In this subsection, we aim to represent non-prioritized operator families as subfamilies of versatile operators.
To do so, we use these relations to model credibility sets of CL revisions, retractable sets of shielded contractions, allowed sets of filtered revisions, and the credible functions used in moderated revision.
Nevertheless, reproducing the whole tree of subfamilies of each of these operators in the AWS framework is beyond the scope of this work.

We start by considering some basic properties that characterize these relations.
\begin{defn}\label{def:cred-quest-relations-properties}
Given $\Cred_w$, $\Quest_w$ two relations over $\Pset(\Interpretations)$, we say:
\begin{itemize}
        \item $\Cred_w$ is \textit{pro-consistent observation} if $A \Cred_w \emptyset$, then $A = \emptyset$;
        \item $\Cred_w$ is \textit{compatible} if $A= \emptyset$ or $A \cap B\neq \emptyset$ then $A\Cred_w B$;
        \item $\Quest_w$ is \textit{questionable:} if $A \Quest_w B$ then $A \not \subseteq B$;
        \item $\Quest_w$ is \textit{normal questionable:} if $A \Quest_w B$ then $A \cap B = \emptyset \neq A$.
\end{itemize}
    A relation $\Cred_w$ is \textit{credible} if it is both pro-consistent observation and compatible.
\end{defn}

These properties are inspired by the work related to credible and retractable sets~\cite{FH1999ShC, H2001}, and can be axiomatically represented through world operators.

\begin{propo}\label{propo:world-change-cred-quest-relations}
    Let $\Cred_w$, $\Quest_w$ be relations from a world operator $\generico_w$. Then:
    \begin{enumerate}
        \item $\Cred_w$ is pro-consistent observation, and it is credible iff $\generico_w$ satisfies:
        \begin{postulate}{WRel}
            \item \label{post:world-credible-vacuity} If $A \cap B \neq \emptyset$ then $(A \generico_w B) \cap B \neq \emptyset$ (world credible vacuity).
        \end{postulate}
        \item $\Quest_w$ is questionable, and it is also normal questionable iff $\generico_w$ satisfies:
        \begin{postulate}{WRel}
                \setcounter{postcounter}{1}
                \item \label{post:world-normal-questionable-vacuity} If $A \cap B \neq \emptyset$ then $(A \generico_w B) \cap A \subseteq B$ (world questionable vacuity).
        \end{postulate}
    \end{enumerate}
\end{propo}

\beginpraax{Proposition}{propo:world-change-cred-quest-relations}
\begin{enumerate}
    \item[] 
    \item $A \Cred_w \emptyset$ means $(A \generico_w \emptyset) \cap \emptyset \neq \emptyset$ or $A = \emptyset$.
    But since $(A \generico_w \emptyset) \cap \emptyset \neq \emptyset$ is false, the only option is 
    $A = \emptyset$.
    Thus $\Cred_w$ is pro-consistent observation.

    Consider the case $A = \emptyset$, then $A \Cred_w B$ by definition and \ref{post:world-credible-vacuity} is trivially satisfied.
    If $A \neq \emptyset$, note that $ A \Cred_w B$ is equivalent to $(A \generico_w B) \cap B \neq \emptyset$.
    Thus, in this context, $\Cred_w$ being compatible means $\generico_w$ satisfying \ref{post:world-credible-vacuity}.
    \item By definition, $A \Quest_w B$ means that $(A \generico_w B) \cap A \not \subseteq B$, which in particular $A \not \subseteq B$. Thus $\Quest_w$ is questionable.

    $\Quest_w$ is normal questionable iff $A \Quest_w B$ implies $A \cap B = \emptyset \neq A$.
    Therefore, if $A \cap B \neq \emptyset$, then $A \not \Quest_w B$.
    By construction, this means $(A \generico_w B) \cap A \cap B^c = \emptyset$, or equivalently $(A \generico_w B) \cap A \subseteq B$.
    Suppose now \ref{post:world-normal-questionable-vacuity} holds, and $A \Quest_w B$.
    By construction, $(A \generico_w B) \cap A \not \subseteq B$, hence $A \cap B = \emptyset$ by \ref{post:world-normal-questionable-vacuity}.
    Also, since $(A \generico_w B) \cap A \cap B^c \neq \emptyset$, we now $A \neq \emptyset$.
\end{enumerate}
\endpraax

A direct corollary of the previous proposition is the characterization of the relations associated with elementary world versatile operators in terms of world-selection functions and their associated elementary world operators.

\begin{cor}\label{cor:vers-cred-quest-relations}
    Let $\vers_w$ be an elementary world versatile constructed by $(\sigma, \rho)$, where $\Cred_w$, $\Quest_w$ are its relations.
    Then $\Cred_w$ is credible, $\Quest_w$ is questionable, and given $A$, $B \subseteq \Interpretations$ each line of the table has equivalent representations:
\begin{center}
\begin{tabular}{cccc}
    \toprule
    \textbf{Relations} & \textbf{world-selection} & \textbf{$\cont_w^\sigma$ and $\cont_w^\rho$} & \textbf{$\rev_w^\sigma$ and $\rev_w^\rho$} \\
    & \textbf{functions $\sigma$ and $\rho$} & \textbf{operators} & \textbf{operators} \\
    \midrule
    $A \Cred_w B$ &
    $(A \cap B) \cup \sigma(A,B^c) \neq \emptyset$ &
    $A \cont_w^\sigma B^c \not \subseteq B^c$ &
    $A \rev_w^\sigma B \neq \emptyset$ \\
    & or $A=\emptyset$ & or $A = \emptyset$ & or $A = \emptyset$ \\
    \addlinespace
    $A \Quest_w B$ &
    $\rho(B,A^c) \neq \emptyset$ &
    $B \cont_w^\rho A^c \not \subseteq B$ &
    $B \rev_w^\rho A \not \subseteq B$ \\
    \addlinespace
    $A \FullQuest_w B$ &
    $\rho(B,A^c) = A \cap B^c \neq \emptyset$ &
    $A \subseteq B \cont_w^\rho A^c \not \subseteq B$ &
    $B \rev_w^\rho A = A \not \subseteq B$ \\
    \bottomrule
\end{tabular}
\end{center}
\end{cor}

Following this result, we say $\sigma$ is the \textit{credible} function and $\rho$ is the \textit{questionable} function of the versatile operator $\vers_w$ constructed by the pair $(\sigma, \rho)$.
Analogously, we refer $\cont_w^\sigma$ and $\rev_w^\sigma$ as the \textit{credible} operators, and $\cont_w^\rho$ and $\rev_w^\rho$ as the \textit{questionable} operators, associated to $\vers$.
These relations allow us to easily confirm that world versatile operators can cover any movement given $A$, $B \subseteq \Interpretations$:
\begin{itemize}
    \item $A \FullQuest_w B$ is associated to either a \textbf{Contraction} movement (if $\sigma(A, B^c) \neq \emptyset$) or a \textbf{No Change} movement (if $\sigma(A, B^c) = \emptyset$);
    \item $A \Cred_w B$ and $A \Quest_w B$ is understood as a \textbf{Moderation} movement;
    \item $A \Cred_w B$ and $A \not \Quest_w B$ means either a \textbf{Revision} (if $\sigma(A, B^c) \neq \emptyset$) or an \textbf{Expansion} movement (if $\sigma(A, B^c) = \emptyset$).
\end{itemize}

We continue by showing that shielded contraction is a subfamily of versatile operators.
From Corollary \ref{cor:rev-cont-as-vers} and \ref{cor:vers-cred-quest-relations}, we can deduce that $\Quest_w$ becomes trivial when the elementary world versatile $\vers_w$ satisfies \ref{post:world-c-inclusion}, since $A \Quest_w B$ iff $A \not \subseteq B$.
Meanwhile, relation $\Cred_w$ represents whether the operator contracts (\textbf{Contraction} movement) or avoids the change (\textbf{No Change} movement), making the elementary world contractions $\cont_w^\sigma$ behave as a shielded contraction.

\begin{thm}\label{thm:world-shielded-contraction-repr}
Let $\cont_w$ be a world operator. The following are equivalent:
\begin{enumerate}
    \item $\cont_w$ satisfies \ref{post:world-c-inclusion} and \ref{post:world-c-recovery};
    \item $\cont_w$ is constructed by an elementary world versatile $\vers_w$ satisfying \ref{post:world-c-inclusion}, where $\Cred_w$ is its associated credible relation, as follows:
    \[
    A \cont_w B = \begin{cases}
        A \vers_w B^c & \si A \Cred_w B^c\\
        A & \cc
    \end{cases}
    \]
    \item $\cont_w$ is constructed as a \textit{world shielded contraction}, i.e., by $\cont'_w$ an elementary world contraction satisfying \ref{post:world-c-success} and $\rel$ a relation over $\Pset(\Interpretations)$, as follows:
    \[
    A \cont_w B = \begin{cases}
        A \cont'_w B & \si A \rel B\\
        A & \cc
    \end{cases}
    \]
\end{enumerate}
\end{thm}

\beginpraax{Theorem}{thm:world-shielded-contraction-repr}
If $\cont_w$ satisfies \ref{post:world-c-inclusion} and \ref{post:world-c-recovery}, then by Theorem \ref{thm:world-contraction-revision-repr-and-identitities} there is a world-selection function $\sigma$ such that $A \cont_w B = A \cup \sigma(A,B)$ for every $A, B \subseteq \Interpretations$.
By Corollary \ref{cor:rev-cont-as-vers}, there is a $\vers_w$ satisfying \ref{post:world-c-inclusion} such that $A \cont_w B = A \vers_w B^c$.
By definition of $\Cred_w$, note that $A \Cred_w B^c$ means that either $A \vers_w B^c \not \subseteq B$ or $A= \emptyset$.
Therefore, $A \not \Cred_w B^c$ is equivalent to $A \neq \emptyset$ and $A \vers_w B^c \subseteq B$.
Since $A \vers_w B^c = A \cont_w B = A \cup \sigma(A,B)$, then $\sigma(A,B) = \emptyset$, otherwise $\sigma(A,B) \not \subseteq B$ by definition.
We conclude that $A \cont_w B = A$.

Given now the $\vers_w$ construction of $\cont_w$, where $\sigma$ is the credible function of $\vers_w$, take the relation 
$ A \rel B$ iff $(A \cap B^c) \cup \sigma(A,B) \neq \emptyset$,
and consider $\overline{\sigma}$ as follows:
    \[
    \overline{\sigma}(A, B) = \begin{cases}
        \sigma(A,B) & \si A \rel B\\
        \sigma'(A,B) & \cc
    \end{cases}
    \]
with $\sigma'(A,B)$ a semantically successful world-selection function.
Take $\cont'_w$ the elementary world contraction constructed by $\overline{\sigma}$.
Note that when $A \rel B$ holds, if $A \subseteq B \subsetneq \Interpretations$, then $A \cap B^c = \emptyset$, hence $\sigma(A,B) \neq \emptyset$.
And since the other case has been selected to make $\overline{\sigma}$ a semantically successful world-selection function, we conclude that $\cont'_w$ satisfies \ref{post:world-c-success} by Theorem \ref{thm:world-cont-rev-classic-repr}.
Moreover, when $A \rel B$, we have that $A \cont'_w B = A \cup \sigma(A,B)$ by construction, and by hypothesis $A \vers_w B^c = A \cup \sigma(A,B) = A \cont_w B$.
Therefore $A \cont'_w B = A \cont_w B$.
If $A \not \rel B$, then $\sigma(A,B) = \emptyset$.
Therefore, $A \cont_w B = A$ again by hypothesis.
We conclude that $\cont_w$ can be constructed as a world shielded contraction.

Finally, assume that $\cont_w$ is constructed as a world shielded contraction. Then, \ref{post:world-c-inclusion} is deduced by the fact that $A$ is a subset of both $A\cont_w B$ and $A$; and \ref{post:world-c-recovery} is deduced by $-_w$ satisfying \ref{post:world-c-recovery} and that $A \cap B \subseteq A$.
\endpraax

Moderated revision \cite{GMR2024} is easier to conceive as a subfamily of versatile operators.
It can be seen as a versatile operator for basic contraction and revision.

\begin{thm}\label{thm:world-mod-repr}
Let $\vers_w$ be an elementary world versatile operator constructed by the pair $(\sigma, \rho)$.
Consider the following postulates for $\vers_w$:
    \begin{postulate}{W$\vers$}
        \setcounter{postcounter}{2}
        \item \label{post:world-credible-consistency} If $B \neq \emptyset$ then $(A \vers_w B) \cap B \neq \emptyset$ (world credible consistency).
        \item \label{post:world-normal-credible-vacuity} If $A \cap B \neq \emptyset$ then $A \vers_w B \subseteq A$ (world normal credible vacuity).
    \end{postulate}
Each line of the table represents equivalent statements for every set $A$, $B \subseteq \Interpretations$:
\begin{center}
\begin{tabular}{cc}
   \toprule
   \textbf{$\vers_w$ postulate} & \textbf{world-selection function property} \\ \midrule
    \ref{post:world-credible-consistency} &
    $\sigma$ is semantically successful
    \\ \addlinespace
    \ref{post:world-normal-credible-vacuity} &
    $\sigma$ is normal
    \\ \addlinespace
    \ref{post:world-normal-questionable-vacuity} &
    $\rho$ is normal
   \\\bottomrule
\end{tabular}
\end{center}
Operators also satisfying these postulates are called world moderated revision.
\end{thm}

\beginpraax{Theorem}{thm:world-mod-repr}
    Given an elementary world versatile $\vers_w$ constructed by $(\sigma, \rho)$, assume that $A \cap B \neq \emptyset$.
    If $\vers_w$ also satisfies \ref{post:world-normal-credible-vacuity}, then it means that $\sigma(A,B^c) = \emptyset$.
    Since $A \cap B \neq \emptyset$ can be seen as $A \not \subseteq B^c$, we conclude that $\sigma$ is a normal world-selection function.
    Analogously, if $\vers_w$ satisfies \ref{post:world-normal-questionable-vacuity}, then it means that $\rho(B,A^c) = \emptyset$.
    Since $A \cap B \neq \emptyset$ can be seen as $B \not \subseteq A^c$, we conclude that $\rho$ is a normal world-selection function.
    Regarding \ref{post:world-credible-consistency}, note that it is equivalent to $\rev_w^\sigma$ satisfying \ref{post:world-r-consistency}, which is the same as $\sigma$ being semantically successful.
\endpraax

Since an operator that satisfies \ref{post:world-r-inclusion}, \ref{post:world-normal-credible-vacuity} and \ref{post:world-normal-questionable-vacuity} it equivalently satisfies \ref{post:world-r-vacuity}, a world moderated revision can be defined by the postulates \ref{post:world-shared-success}, \ref{post:world-credible-consistency} and \ref{post:world-r-vacuity}.

Other non-prioritized operators are more difficult to associate with the versatile family.
For example, shielded contractions and CL revision~\cite{FH1999ShC,H2001} are presented as duals, but via a variation of the Levi Identity.
Nevertheless, we have seen that the dual of world shielded contraction, a.k.a. elementary world contraction, are elementary world revision.
We now show CL revision and filtered revision~\cite{Bonanno2020,garapa_2022}, a generalization of both CL revision and shielded contraction, are in fact other non-prioritized operators that lie between elementary world contractions and revisions.
But the key difference from elementary versatile operators is that $\Cred_w$, $\Quest_w$, and $\FullQuest_w$ are connected.

\begin{defn}\label{def:cred-quest-entanglement}
    Let $\generico_w$ be a world operator and $\Cred_w$, $\Quest_w$, $\FullQuest_w$  its relations.
    \begin{description}
        \item[The relation $\Quest_w$ is \textbf{full}] when $\Quest_w$ and $\FullQuest_w$ are equivalent.
        \item[The pair $(\Cred_w , \Quest_w)$ is \textbf{entangled}] if $A \not \Cred_w B$ then $A \Quest_w B$;
        \item[The pair $(\Cred_w , \Quest_w)$ is \textbf{strongly entangled}] when $\not \Cred_w$ and $\Quest_w$ are equivalent;
    \end{description}
\end{defn}

The entanglement situation of $\Cred_w$ and $\Quest_w$ could be interpreted as ``every non-credible set is a questionable set''.
Strengthening this position is to assume that these concepts are equivalent.
Another situation is to force every \textbf{Moderation} movement into a \textbf{Contraction} movement, represented by $\Quest_w$ being full.

Connecting relations of world versatile operators means connecting also their world-selection functions.
However, these connections are directly deduced by Corollary \ref{cor:vers-cred-quest-relations}.
Thus, we focus on how these properties relate to postulates.

\begin{propo}\label{propo:world-connections-cred-quest-relations}
    Let $\generico_w$ be an elementary world versatile operator, and $\Cred_w$, $\Quest_w$, $\FullQuest_w$ its relations.
    Consider the following postulates:
    \begin{postulate}{WRel}
        \setcounter{postcounter}{2}
        \item \label{post:world-full-knowledge-preservation} If $(A \vers_w B) \cap A \not \subseteq B$ then $A \subseteq A \vers_w B$ (world knowledge preservation).
        \item \label{post:world-consistency-preservation} If $A \neq \emptyset$ then $A \vers_w B \neq \emptyset$ (world consistency preservation).
        \item \label{post:world-weak-relative-success} Either $A \vers_w B \subseteq B$ or $A \vers_w B \subseteq A$ (world weak relative success).
    \end{postulate}
    Then, the following equivalences hold:
    \begin{enumerate}
        \item $\Quest_w$ is full iff $\generico_w$ satisfies \ref{post:world-full-knowledge-preservation};
        \item $(\Cred_w , \Quest_w)$ is entangled iff $\generico_w$ satisfies \ref{post:world-consistency-preservation};
        \item $(\Cred_w , \Quest_w)$ is strongly entangled iff $\generico_w$ satisfies \ref{post:world-normal-questionable-vacuity}, \ref{post:world-consistency-preservation}, \ref{post:world-weak-relative-success}.
    \end{enumerate}
\end{propo}

\beginpraax{Proposition}{propo:world-connections-cred-quest-relations}
Relation $\Quest_w$ being full is equivalent to $\rho(B,A^c)$ being either $A \cap B^c$ or $\emptyset$, which is equivalent to say $\rho(B,A^c) \neq \emptyset$ implies $\rho(B,A^c) = A \cap B^c$.
By applying Corollary \ref{cor:vers-cred-quest-relations}, this implication is equivalent to \ref{post:world-full-knowledge-preservation}.

When the pair $(\Cred_w, \Quest_w)$ is entangled, one of these three situations holds:
\[
(A \cap B) \cup \sigma(A,B^c) \neq \emptyset \quad , \quad A = \emptyset \quad , \quad \rho(B,A^c)\neq \emptyset
\]
Hence, $A \neq \emptyset$ implies $(A \cap B) \cup \sigma(A,B^c) \cup \rho(B,A^c)\neq \emptyset$, i.e. $\vers_w$ satisfies \ref{post:world-consistency-preservation}.

If the pair $(\Cred_w, \Quest_w)$ is strongly entangled, we already know \ref{post:world-consistency-preservation} holds.
Also, if $A \cap B \neq \emptyset$, then $A \Cred_w B$, which means that $A \not \Quest_w B$.
By definition of $\Quest_w$, this is $(A \vers_w B) \cap A \subseteq B$.
Hence, $\vers_w$ satisfies \ref{post:world-normal-questionable-vacuity}.
Assume now $A \vers_w B \not \subseteq B$, this only happens when $\rho(B,A^c) \neq \emptyset$.
By Corollary \ref{cor:vers-cred-quest-relations}, we have $A \Quest_w B$, and since $(\Cred_w, \Quest_w)$ is strongly entangled, it implies that $A \not \Cred_w B$.
Once again by Corollary \ref{cor:vers-cred-quest-relations}, $(A \cap B) \cup \sigma(A,B^c)$, which means that $A \vers_w B = \rho(B,A^c)$.
Therefore, $A \vers_w B \subseteq A$ by definition of world-selection function.

Suppose now $\vers_w$ satisfies \ref{post:world-normal-questionable-vacuity}, \ref{post:world-consistency-preservation}, \ref{post:world-weak-relative-success}, and consider the case where $A \Quest_w B$.
By definition of $\Quest_w$ and \ref{post:world-weak-relative-success}, we have that $A \vers_w B \subseteq A$.
This implies that $\sigma(A,B^c) = \emptyset$, and by \ref{post:world-normal-questionable-vacuity}, we have that $A \cap B = \emptyset$.
Therefore, $(A \cap B) \cup \sigma(A,B^c) = \emptyset$, i.e. $A \not \Cred_w B$.
If $A \not \Quest_w B$, then $\rho(B, A^c) = \emptyset$ by Corollary \ref{cor:vers-cred-quest-relations}.
Hence, $A \vers_w B = (A \cap B) \cup \sigma(A,B^c)$, and by \ref{post:world-normal-questionable-vacuity}, we have that either $(A \cap B) \cup \sigma(A,B^c) \neq \emptyset$ or $A = \emptyset$.
Therefore, $A \Cred_w B$.
\endpraax

These postulates allow us to characterize the remaining non-prioritized operators as subfamilies of versatile operators.
We start with world filtered revision.

\begin{thm}\label{thm:elementary-world-filtered-revision-repr}
Let $\vers_w$ be a world operator. The following are equivalent:
\begin{enumerate}
    \item $\vers_w$ is an elementary world versatile operator satisfying \ref{post:world-full-knowledge-preservation}.
    \item If $\Cred_w$ and $\Quest_w$ are the associated relations of $\vers_w$, there is an elementary world revision $\rev_w$ such that:
    \[
    A \vers_w B = \begin{cases}
        A \rev_w B & \si A \not \Quest_w B\\
        (A \rev_w B) \cup A & \si A \Cred_w B \mbox{ and } A \Quest_w B\\
        A & \si A \not \Cred_w B \mbox{ and } A \Quest_w B\\
    \end{cases}
    \]
    \item $\vers_w$ is constructed as an \textit{elementary world filtered revision}, i.e., by an elementary world revision $\rev_w$, and two disjoint relations $\rel_{cred}$, $\rel_{allow}$, as follows:
    \[
    A \vers_w B = \begin{cases}
        A \rev_w B & \si A \rel_{cred} B\\
        (A \rev_w B) \cup A & \si A \rel_{allow} B\\
        A & \cc\\
    \end{cases}
    \]
\end{enumerate}
\end{thm}

\beginpraax{Theorem}{thm:elementary-world-filtered-revision-repr}
If $\vers_w$ is an elementary world versatile operator constructed by $(\sigma, \rho)$, recall that $A \vers_w B = A \rev_w^\sigma B$ when $A \not \Quest_w B$.
Moreover, by \ref{post:world-full-knowledge-preservation}, we have that $A \vers_w B = (A \rev_w^\sigma B) \cup A$ when $A \Quest_w B$.
Lastly, note $A \vers_w B = A$ when $A \not \Cred_w B$ and $A \Quest_w B$, then.
Then, it is enough to take $\rev_w$ as $\rev_w^\sigma$. 

If $\vers_w$ is constructed by its relations and an elementary world revision, consider $\rel_{cred}$ as $\not \Quest_w$, and $\rel_{cred}$ as the intersection of $\Cred$ and $\Quest$ to construct it as an elementary world filtered revision.

Lastly, we show that the elementary world filtered revision construction satisfies the postulates.
Both postulates \ref{post:world-shared-success} and \ref{post:world-r-inclusion} are satisfied by construction.
If $(A \vers_w B) \cap A \not \subseteq B$, then $A \vers_w B \not \subseteq B$.
This means $A \vers_w B$ is either $(A \rev_w B) \cup A$ or $A$.
In any case, $A \subseteq A \vers_w B$.
Therefore, $\vers_w$ satisfies \ref{post:world-full-knowledge-preservation}.
\endpraax

As stated before, the family of world CL revision is based on a stronger connection between $\Cred_w$ and $\Quest_w$ relations, by adding postulate \ref{post:world-weak-relative-success}.

\begin{thm}\label{thm:credibility-limited-revision-repr}
Let $\vers_w$ be a world operator. The following are equivalent:
\begin{enumerate}
    \item $\vers_w$ is an elementary world versatile operator satisfying \ref{post:world-full-knowledge-preservation}, \ref{post:world-weak-relative-success}.
    \item if $\Quest_w$ is the associated questionable relations of $\vers_w$, there is an elementary world revision $\rev_w$ such that:
    \[
    A \vers_w B = \begin{cases}
        A \rev_w B & \si A \not \Quest_w B\\
        A & \si A \Quest_w B\\
    \end{cases}
    \]
    \item $\vers_w$ is constructed as an \textit{elementary world CL revision}, i.e., by $\rev_w$ an elementary world revision, and $\rel_{cred}$ a relation over $\Pset(\Interpretations)$, as follows:
    \[
    A \vers_w B = \begin{cases}
        A \rev_w B & \si A \rel_{cred} B\\
        A & \cc\\
    \end{cases}
    \]
\end{enumerate}
\end{thm}

\beginpraax{Theorem}{thm:credibility-limited-revision-repr}
By Theorem \ref{thm:elementary-world-filtered-revision-repr}, we only need to show $A \Quest_w B$ implies $A \vers_w B = A$ when \ref{post:world-weak-relative-success} holds.
Assume $\vers_w$ satisfies \ref{post:world-weak-relative-success} and $A \Quest_w B$.
This means $(A \vers_w B) \cap A \not \subseteq B$ by definition of $\Quest_w$.
Therefore $A \vers_w B \not \subseteq B$ and $A \subseteq A \vers_w B$ by \ref{post:world-full-knowledge-preservation}.
Hence, $A \vers_w B = A$ by \ref{post:world-weak-relative-success}.
If $\vers_w$ is constructed by $\Quest_w$ and an elementary world revision, consider $\rel_{cred}$ as $\not \Quest_w$ to construct it as an elementary world CL revision.
Suppose now $\vers_w$ is constructed as an elementary world CL revision.
By construction, it satisfies \ref{post:world-weak-relative-success}, and since it is an elementary world filtered revision, it satisfies \ref{post:world-shared-success}, \ref{post:world-r-inclusion}, and \ref{post:world-full-knowledge-preservation}.
\endpraax

Note any operator satisfying \ref{post:world-full-knowledge-preservation} and \ref{post:world-weak-relative-success} equivalently satisfies:
\begin{postulate}{W$\clr$}
    \item \label{post:world-relative-success} Either $A \vers_w B \subseteq B$ or $A \vers_w B = A$ (world relative success).
\end{postulate}
This is the postulate that typically represents CL revisions.
Also, since \ref{post:world-weak-relative-success} implies \ref{post:world-shared-success}, an these operators can be defined by \ref{post:world-relative-success} and \ref{post:world-r-inclusion}.

A CL revision that uses $\Cred_w$ instead of $\not \Quest_w$ means its relations are strongly entangled, thus it satisfies \ref{post:world-normal-questionable-vacuity} and \ref{post:world-consistency-preservation}.
If it also satisfy \ref{post:world-normal-credible-vacuity}, we have normal world CL revision~\cite{GMR2024}, which can be defined by \ref{post:world-relative-success}, \ref{post:world-r-vacuity} and \ref{post:world-consistency-preservation}.
This last theorem affirms our definition of moderated revision matches the original construction~\cite{GMR2024}, which uses normal CL revision.

\begin{thm}\label{thm:classic-revision-repr}
Let $\rev_w$ be a world operator. The following are equivalent:
\begin{enumerate}
    \item $\rev_w$ is an elementary world revision satisfying \ref{post:world-r-vacuity};
    \item there is a normal CL revision $\vers_w$ such that $A \rev_w B = (A \vers_w B) \cap B$.
\end{enumerate}    
Moreover, there is a one-to-one relation between these operators.
\end{thm}

\beginpraax{Theorem}{thm:classic-revision-repr}
    Given $\rev_w$ an elementary world revision satisfying \ref{post:world-r-vacuity}, define:
    \[
    A \vers_w B = \begin{cases}
        A \rev_w B & \si (A \rev_w B) \cap B \neq \emptyset\\
        A & \cc
    \end{cases}
    \]
    By construction $A \vers_w B$ satisfies \ref{post:world-relative-success}, since $A \rev_w B \subseteq B$, and \ref{post:world-r-vacuity} through $\rev_w$.
    Also by construction, $A \vers_w B \neq \emptyset$ when $A \neq \emptyset$.
    Thus, it satisfies \ref{post:world-consistency-preservation}.

    If $\vers_w$ is a normal CL revision such that $A \rev_w B = (A \vers_w B) \cap B$, then $\rev_w$ satisfies \ref{post:world-r-success} by construction, and \ref{post:world-r-vacuity} through $\vers_w$.
    Since \ref{post:world-r-vacuity} implies \ref{post:world-r-inclusion}, $\rev_w$ is an elementary world revision.

    The relation $A \rev_w B = (A \vers_w B) \cap B$ defines a surjective function from $\vers_w$ to $\rev_w$.
    Consider then two normal CL revision $\vers_w$ and $\vers_w'$ such that they are related to the same $\rev_w$, i.e. $(A \vers_w B) \cap B = A \rev_w B = (A \vers_w' B) \cap B$.
    If $A \cap B \neq \emptyset$, we know $A \vers_w B = A \vers_w' B = A \cap B$ by \ref{post:world-r-vacuity}.
    If $A \cap B = \emptyset$, consider the case where $A \vers_w B = A \neq \emptyset$.
    By hypothesis, $(A \vers_w' B) \cap B = \emptyset$, thus $A \vers_w' B \subseteq B^c$.
    Since $A \neq \emptyset$, we know $A \vers_w' B \neq \emptyset$ by \ref{post:world-consistency-preservation}.
    Therefore $A \vers_w' B = A$ by \ref{post:world-relative-success}.
    Assume now $\emptyset \neq A \vers_w B \subseteq B$.
    Since $A \cap B = \emptyset$, we deduce $A \vers_w B \subseteq A^c$.
    Therefore, $\emptyset \neq (A \vers_w' B) \cap B \subseteq A^c$.
    We deduce $A \vers_w' B \subseteq B$ by \ref{post:world-relative-success}, hence $A \vers_w B = A \vers_w' B$.
    Lastly, if $A \vers_w B = \emptyset$, then $A = \emptyset$ by \ref{post:world-consistency-preservation} and $(A \vers_w' B) \cap B = \emptyset$ by hypothesis. Then, either $A \vers_w' B \subseteq B^c \cap B$ or $A \vers_w' B \subseteq A$.
    In any case, $A \vers_w' B = \emptyset$.
    We conclude that $\vers_w$ and $\vers_w'$ are the same operator, implying that the function from $\vers_w$ to $\rev_w$ is also injective, and therefore, bijective.
\endpraax

With this, we finish the presentation of both non-prioritized and classical operators in the AWS framework as part of the world versatile family.

\section{AWS applied to Tarskian Logics}\label{sec:AWS-applied-TLogics} %%%%%%%%%%%%%%%%%%%%%%%%%%%%%%%%%%%%%%

Here we present the tools needed to recover the classical frameworks and their operators from AWS and world versatile operators.
Although our main focus is on CPL logics, the ideas presented here allow us to apply our framework to any Tarskian logic.
These are a combination of some known results, such as Suszko's Reduction~\cite{Suszko1977TheFA, Caleiro2003SuszkosTA}, Galois Connections~\cite{Lindstrom2022}, and Stone Dualities~\cite{Stone1936}.
Therefore, the novelty here is in how these tools develop into an implementation of AWS in a particular set of ``well-behaved'' Tarskian logics, where CPL is included.
From now on, the language of a logic is an unstructured set, unless stated otherwise, focusing on the consequence operator.

\begin{defn}\label{def:tasrkian-logic}
    Given a language $\Lang$ and a function $Cn:\Pset(\Lang) \rightarrow \Pset(\Lang)$ called consequence operator, we say $\Logic = \langle \Lang , Cn \rangle$ is a Tarskian logic if $Cn$ satisfies:
    \begin{postulate}{$T$}
        \item $M \subseteq Cn(M)$ (inclusion);
        \item if $M \subseteq N$ then $Cn(M) \subseteq Cn(N)$ (monotony);
        \item $Cn(Cn(M)) \subseteq Cn(M)$ (idempotence);
    \end{postulate}
\end{defn}

Suszko's Reduction affirms that every Tarskian logic can be represented by a subset $\Valuations$ of $\{0,1\}^\Lang = \{ f: \Lang \rightarrow \{0,1\} \}$.
We call these functions valuations of $\Logic$.

\begin{thm}[Suszko's Reduction]\label{thm:Tarskian-suszko-reduction}
    Given a Tarskian logic $\Logic = \langle \Lang , Cn \rangle$, there is a subset $\Valuations$ of $\{0,1\}^\Lang$ and a function $\PLtoPV : \Pset(\Lang) \rightarrow \Pset(\Valuations)$ defined as $\PLtoPV[M] = \{ v \in \Valuations \tq M \subseteq v^{-1}(\{1\}) \}$ such that:
    \[
    Cn(M) = \bigcap \limits_{v \in \PLtoPV[M]} v^{-1}(\{1\})
    \]
\end{thm}

\beginpraax{Theorem}{thm:Tarskian-suszko-reduction}
    Consider $\BeliefSets=\{K \subseteq \Pset(\Lang) \tq K = Cn(K) \}$.
    Given $M \subseteq \Lang$, $Cn(M)$ is the smallest element of $\BeliefSets$ containing $M$: $M \subseteq Cn(M)$ by inclusion, and if $M \subseteq K \subseteq Cn(M)$ then $Cn(M) \subseteq Cn(K) \subseteq Cn(Cn(M))$ by monotony, thus $Cn(M) = K$ by idempotence.
    By taking $v_K$ the characteristic functions of every $K \in \BeliefSets$, i.e. $v_K^{-1}(\{1\}) = K$, hence
    $
    \bigcap \limits_{M \subseteq v_K^{-1}(\{1\})} v_K^{-1}(\{1\}) = \bigcap \limits_{M \subseteq K} K = Cn(M)
    $.
\endpraax

The set $\Valuations$ can be understood as a way of representing the elements of the image of $Cn$ as intersections of specific sets associated with $\Valuations$.
However, in Theorem \ref{thm:Tarskian-suszko-reduction}, some valuations may be redundant.
For example $v_\Lang$, the function that returns $1$ for every $\psi \in \Lang$, is irrelevant since $v_\Lang^{-1}(\{1\}) = \Lang$, and we can represent $\Lang$ as the intersection of the empty set.
We formalize this construction.

\begin{defn}\label{def:functions-and-sets-valuation-logics}
    We call $\LogRepr = \langle \Lang, \Valuations \rangle$ a Tarskian representation, where $\Lang$ is a language and $\Valuations$ a subset of $\{0,1\}^\Lang$.
    Every $v \in \Valuations$ is known as a (logical) valuation, and the constant function $\mathbf{1}$ is never considered.
    We note $T_v = v^{-1}(\{1\})$, i.e., $v$ is the characteristic function of the set $T_v$.
    Every $\LogRepr$ is associated with the following functions:
    \begin{itemize}
        \item $\PLtoPV: \Pset(\Lang) \longrightarrow \Pset(\Valuations)$ such that $\PLtoPV[M] = \{ v \in \Valuations \tq M \subseteq T_v \}$;
        \item $\PVtoPL: \Pset(\Valuations) \longrightarrow \Pset(\Lang)$ such that $\PVtoPL(A) = \bigcap \limits_{v \in A} T_v$;
        \item $Cn\evdot= \PVtoPL(\PLtoPV)$ and $Cl\evdot= \PLtoPV[\PVtoPL\evdot]$. \\
    We say $\LogRepr$ represents the logic $\Logic = \langle \Lang , Cn \rangle$.
    Suszko's Reduction affirms that there is always a Tarskian representation for a given Tarskian logic.
    \end{itemize}
\end{defn}

Lindström~\cite{Lindstrom2022} shows that the pair $(\PLtoPV, \PVtoPL)$ is a Galois Connection, giving us a way to link $\Pset(\Lang)$ and $\Pset(\Valuations)$, and showing that every Tarskian representation $\LogRepr$ represents a Tarskian logic.
These results are also related to the one presented in Grove~\cite{Grove1988}, but this generalization avoids the situation that appeared by considering an infinite context~\cite{PriestTanaka1997}.

\begin{thm}\label{thm:galois-connection}
    Let $\LogRepr = \langle \Lang, \Valuations \rangle$ be a Tarskian representation.
    Then the pair $(\PLtoPV, \PVtoPL)$ is an antitone Galois connection\footnote{Lindström~\cite{Lindstrom2022} works with an equivalent definition of Galois connection.}, i.e., given $A \in \Pset(\Valuations)$ and $M \in \Pset(\Lang)$:
    \[
    A \subseteq \PLtoPV[M] \qquad \ssi \qquad M \subseteq \PVtoPL(A)
    \]
    Also, given $S \subseteq \Pset(\Valuations)$ and $X \subseteq \Pset(\Lang)$, and defining $\PVtoPL(S) = \{ \PVtoPL(A) \tq A \in S \}$ and $\PLtoPV[X] = \{ \PLtoPV[M] \tq M \in X \}$, the following properties are satisfied:
    \begin{enumerate}
        \item $\PLtoPV$ and $\PVtoPL$ are antitone (i.e., they invert the inclusion order);
        \item $Cn$ and $Cl$ are Tarskian operators, and $Cl(\emptyset) = \emptyset$;
        \item $\PLtoPV[M] = \PLtoPV[ Cn(M) ] = Cl(\PLtoPV[M])$ and $\PVtoPL(A) = Cn(\PVtoPL(A)) = \PVtoPL(Cl(A))$;
        \item $\PLtoPV[\bigcup X] = \bigcap \PLtoPV[X]$ and $\bigcup \PLtoPV[X] \subseteq \PLtoPV[ \bigcap X ]$;
        \item $\PVtoPL \left( \bigcup S \right) = \bigcap \PVtoPL (S)$ and $\bigcup \PVtoPL(S) \subseteq \PVtoPL\left( \bigcap S \right)$;
    \end{enumerate}
\end{thm}

\beginpraax{Theorem}{thm:galois-connection}
    Let us first check $(\PLtoPV, \PVtoPL)$ is an antitone Galois connection:
    \[
    A \subseteq \PLtoPV[M] \quad \mbox{ iff } \quad M \subseteq T_v \mbox{ for every } v \in A \quad \mbox{ iff } \quad M \subseteq \bigcap \limits_ {v \in A} T_v = \PVtoPL(A)
    \]
    
    Now we continue with the list of properties:
    \begin{enumerate}
        \item If $M \subseteq N$, given a $v \in \|N\|$ we have $N \subseteq T_v$.
        Hence, $M \subseteq T_v$, i.e. $v \in \PLtoPV[M]$.
        Therefore, $\|N \| \subseteq \PLtoPV[M]$.
        If $A \subseteq B$, given $v \in A$ we have $\PVtoPL(B) \subseteq T_v$.
        Therefore, $\PVtoPL(B) \subseteq \bigcap \limits_ {v \in A} T_v = \PVtoPL(A)$.
        
        \item Since $M \subseteq T_v$ for every $v \in \PLtoPV[M]$, we can deduce that $M \subseteq Cn(M)$ by definition of $Cn$.
        Thus $Cn$ satisfies inclusion.
        Monotony is directly deduced by $\PLtoPV$ and $\PVtoPL$ being antitone.
        By the definition of $Cn(M)$, we deduce that if $v \in \PLtoPV[M]$, then $Cn(M) \subseteq T_v$.
        Therefore, $\PLtoPV[M] \subseteq \| Cn(M) \|$.
        And by inclusion and $\PLtoPV$ being antitone, we have $\|Cn(M) \| \subseteq \PLtoPV[M]$.
        Thus $\PLtoPV[M] = \| Cn(M) \|$ by the double inclusion.
        Applying $\PVtoPL$, we have that $Cn(Cn(M)) = Cn(M)$.
        We conclude that $Cn$ is a Tarskian operator.
        
        If $v \in A$ then $\PVtoPL(A) \subseteq T_v$ by the Galois Connection.
        Since $\PLtoPV$ is antitone, $\| T_v \| \subseteq Cl(A)$.
        We also know $v \in \| T_v \|$, therefore $v \in Cl(A)$.
        Thus $A \subseteq Cl(A)$.
        Monotony is directly deduced by $\PLtoPV$ and $\PVtoPL$ being antitone.
        For idempotence, recall that $\PLtoPV[M] = \| Cn(M) \|$.
        By taking $M = \PVtoPL(A)$, we deduce $Cl(A) = \PLtoPV[Cn(\PVtoPL(A))] = Cl(Cl(A))$.
        Thus $Cl$ is a Tarskian operator.
        Lastly, recall that given $v \in \Valuations$ there is a $\psi \in \Lang$ such that $v(\psi)=0$.
        Hence, $Cl(\emptyset) = \| \PVtoPL(\emptyset) \| = \| \Lang \| = \emptyset$ by definition of $Cl$.
        
        \item Since $\PLtoPV[M] = \|Cn(M)\|$ and $\PVtoPL(A) = \PVtoPL(Cl(A))$, by definition of $Cn$ and $Cl$ it is directly deduced $\| Cn(M) \| = Cl( \PLtoPV[M])$ and $\PVtoPL(Cl(A)) = Cn(\PVtoPL(A))$.

        \item Note that $v \in \bigcap \PLtoPV[X]$ is equivalent to $M \subseteq T_v \mbox{ for every } M \in X$, which means $\bigcup X \subseteq T_v$, which is the same as $v \in \PLtoPV[ \bigcup X ]$.
        Also, if $v \in \bigcup \PLtoPV[X]$, there is $M_v \in X$ such that $v \in \|M_v\|$.
        Therefore $\bigcap X \subseteq T_v$, i.e. $v \in \| \bigcap X \|$.
        
        \item Note that $\bigcap \PVtoPL(S) = \bigcap \limits_{A \in S} \left(\bigcap \limits_ {v \in A} T_v\right) = \bigcap \limits_{v \in \bigcup S} T_v = \PVtoPL \left( \bigcup S \right)$.
        Also, if $v \in \bigcap S$ then $\PVtoPL(A) \subseteq T_v$ for every $A \in S$.
        Therefore, $\bigcup \limits_{A \in S} \PVtoPL(A) \subseteq T_v$.
        Since this holds for every $v \in \bigcap S$, by definition of $\PVtoPL$, we have $\bigcup \limits_{A \in S} \PVtoPL(A) \subseteq \PVtoPL (\bigcap S)$.
    \end{enumerate}
\endpraax

From the infinite case~\cite{PriestTanaka1997}, we know $\PLtoPV$ and $\PVtoPL$ are not in general bijections.
Nevertheless, some bijections arise when restricting the domain of these functions.

\begin{propo}\label{propo:sets-bijections}
    Let $\Logic = \langle \Lang, Cn \rangle$ be a Tarskian logic and $\LogRepr = \langle \Lang, \Valuations \rangle$ its Tarskian representation. 
    Then, $\PVtoPL$ and $\PLtoPV$ define a bijection between:
    \begin{enumerate}
        \item the set of closed sets of $\Valuations$, $\ClosedSets = \lbrace A \subseteq \Valuations \, | \, A = Cl( A ) \rbrace$, and the set of theories of $\Logic$, also known as belief sets, $\BeliefSets = \lbrace K \subseteq \Lang \, | \, K = Cn( K ) \rbrace$;
        \item the set of closed singletons of $\Valuations$, $\{v \in \Valuations \tq Cl(\{v\}) = \{v\} \}$, and the set of maximally consistent sets, $\MaxConsSets = \lbrace T \in \BeliefSets \setminus \Lang \tq \forall \alpha$, $Cn(T \cup \{ \alpha \}) = T$ or $\Lang \rbrace$;
        \item the set of clopen sets, $\ClopenSets = \lbrace A \subseteq \Valuations \tq A, A^c \in \ClosedSets \rbrace$, and the set of complemented theories, $\BeliefSets_f = \lbrace K \in \BeliefSets \tq Cn(K \cup \PVtoPL(\PLtoPV[K]^c)) = \Lang \rbrace$.
    \end{enumerate}
\end{propo}

\beginpraax{Proposition}{propo:sets-bijections}
Recall that given $A \subseteq \Valuations$, we have $\PVtoPL(A) = \PVtoPL(Cl(A)) = Cn(\PVtoPL(A))$.
Analogously $\PLtoPV[M] = \PLtoPV[Cn(M)] = Cl(\PLtoPV[M])$ for every $M \subseteq \Lang$.
Thus, the image of $\PLtoPV$ is $\ClosedSets$ and the image of $\PVtoPL$ is $\BeliefSets$.
Therefore, these functions are surjective when restricted to $\BeliefSets$ and $\ClosedSets$.
Also, if $A$, $B \in \ClosedSets$ such that $\PVtoPL(A) = \PVtoPL(B)$, by applying $\PLtoPV$, we have $Cl(A) = Cl(B)$, i.e., $A=B$.
And if $M$, $N \in \BeliefSets$ such that $\PLtoPV[M] = \|N\|$, by applying $\PVtoPL$, we have $Cn(M) = Cn(N)$, i.e. $M=N$.
This means these functions are injective when restricted to $\BeliefSets$ and $\ClosedSets$.
Lastly, $Cn$ and $Cl$, when restricted to $\BeliefSets$ and $\ClosedSets$, are the identity functions of these sets.
Thus, $\PVtoPL$ and $\PLtoPV$ define a bijection between $\BeliefSets$ and $\ClosedSets$.

We use the previous bijection to prove the others.
Given $v \in \Valuations$, note that $\PVtoPL(\{v\}) = T_v$ and, therefore, $T_v \in \BeliefSets$.
If $Cl(\{v\}) = \{v\} $, then there is no $v' \neq v$ such that $T_v \subseteq T_{v'}$.
If $T_v \not \in \MaxConsSets$, then there is $\alpha \not \in T_v$ such that $Cn(T_v \cup \alpha) \neq \Lang$.
By applying $\PLtoPV$ and its properties, $\emptyset \neq \| T_v \cup \{\alpha\} \| = \| T_v \| \cap \| \{\alpha\} \| = \{v\} \cap \| \{\alpha\} \|$. Thus $\alpha \in T_v$, which leads to a contradiction.
Therefore $T_v \in \MaxConsSets$.
If $T \in \MaxConsSets$, then $\|T\| \neq \emptyset$ by definition, since $\Lang \not \in \MaxConsSets$.
Consider $v \in \|T\|$, hence $T \subseteq T_v$.
Note that $T_v \neq \Lang$.
If the inclusion is strict, then there is a formula $\alpha \not \in T$ such that $\alpha \in T_v$.
Hence, $T \cup \{\alpha\} \subseteq T_v$.
By monotony and idempotence of $Cn$, and the definition of $T$, $\Lang = Cn(T \cup \{\alpha\}) \subseteq T_v \neq \Lang$, leading to a contradiction.
We conclude that $T = T_v$.
Thus $\|T \| = \{v\}$ and $Cl(\{v\})=\{v\}$.
    
If $K \in \BeliefSets_f$ then $\PLtoPV[ Cn(K \cup \PVtoPL(\PLtoPV[K]^c) ]= \emptyset$.
Hence, by applying Theorem \ref{thm:galois-connection}, 
$\emptyset = \PLtoPV[ Cn(K \cup \PVtoPL(\PLtoPV[K]^c) ] = \PLtoPV[ K \cup \PVtoPL(\PLtoPV[K]^c ] = \PLtoPV[K] \cap Cl(\PLtoPV[K]^c)$.
Since we also know that $\PLtoPV[K]^c \subseteq Cl(\PLtoPV[K]^c)$, we have that $Cl(\PLtoPV[K]^c) = \PLtoPV[K]^c$, i.e. $\PLtoPV[K] \in \ClopenSets$. 
Assume now that $A \in \ClopenSets$, then $\emptyset = A \cap A^c = Cl(A) \cap Cl(A^c) = \| \PVtoPL(A) \cup \PVtoPL(A^c) \|$
again by Theorem \ref{thm:galois-connection}. Thus $Cn(\PVtoPL(A) \cup \PVtoPL(A^c)) = \Lang$, i.e., $\PVtoPL(A) \in \BeliefSets_f$.
\endpraax

This result is the groundwork where our proposal connects to Tarskian logics, by extending the bijections $\PLtoPV$ and $\PVtoPL$ to lattice isomorphisms between $\BeliefSets$ and $\ClosedSets$.

%%%%%%%%%%%%%%%%%%%%%%%%%%%%%%%%%%%%%%%%%%%%%%%%%%%%%%%%%%%%%%%%%%%%%%%%%%%%%%%%%%%%%%%%%%%%%%
\subsection{Belief set operators}

By considering $\Interpretations$ of Section \ref{sec:abstract-world-semantic} as the index set of $\Valuations$, we can move from $\Logic$ and our proposal using $\PLtoPV$ and $\PVtoPL$.
The bijections from Proposition \ref{propo:sets-bijections} lead us to associate multiple change operators restricted to belief sets with world operators restricted to closed sets.

\begin{defn}\label{def:cpl-world-based-operator}
    Given a Tarskian logic, we call $\generico: \BeliefSets \times \BeliefSets \rightarrow \BeliefSets$ a belief set operator.
    We say $\generico$ is world-based, or equivalently that is based on $\generico_w$, if there is a world operator $\generico_w$ such that $ \PLtoPV[K \generico M] = \PLtoPV[K] \generico_w \PLtoPV[M] $ for every $K$, $M\in \BeliefSets$.
    We say a world operator $\generico_w$ is closed if $A \generico_w B \in \ClosedSets$ for every $A$, $B \in \ClosedSets$.
\end{defn}

This definition imposes a constraint from the AWS framework perspective: only closed world operators relate to world-based operators.
More precisely, it implies that every belief set operator is a world-based operator, defining a partition over closed world operators.

\begin{cor}\label{cor:bs-ops-are-world-based}
    Let $\Logic$ be a Tarskian logic with a Tarskian representation $\langle \Lang, \Valuations \rangle$, $Op(\BeliefSets)$ the set of all the belief set operators of $\Logic$, and $Op_{\ClosedSets}(\Valuations)$ the set of all the closed world operators.
    We say $\generico_w$, $\circ_w \in Op_{\ClosedSets}(\Valuations)$ are closed-related, noted as $\generico_w \equiv_{\ClosedSets} \circ_w$, if $A \generico_w B = A \circ_w B$ for every $A$, $B \in \ClosedSets$.
    Then $\equiv_{\ClosedSets}$ is an equivalence relation, and the world-based correspondence induced from Definition \ref{def:cpl-world-based-operator} is a bijection between $Op(\BeliefSets)$ and the class of equivalencies defined from $\equiv_{\ClosedSets}$.
\end{cor}

Thus, belief set operators are not only world-based but also represent a class of equivalence in the set of closed world operators.
It also allows us to understand closed world operators as, essentially, world operators restricted to $\ClosedSets$.

So far, we have established several connections between $\BeliefSets$ and $\ClosedSets$.
We now extend these connections to their underlying lattice structures.
Using first order logic over the signature of lattices as a bridge, we make explicit the equivalence, up to isomorphism, between axiomatically characterized belief set operators and axiomatically characterized world operators restricted to closed sets.

\begin{rem}\label{rem:galois-connection-extended-to-postulates}
    Let $\Logic$ be a Tarskian logic with a Tarskian representation $\langle \Lang, \Valuations \rangle$.
    Given $\BeliefSets$, consider $\cap$ the arbitrary intersection as the meet operator, $\expn$ the closure of the arbitrary union of $\Pset(\Lang)$ as the join operator, and $Cn(\emptyset)$, $\Lang$ are the bottom and top elements, respectively.
    Analogously, given $\ClosedSets$, consider $\expn_w$ the closure of arbitrary union as the meet operator and $\cap_w$ the arbitrary intersection of $\Pset(\Valuations)$ as the join operator, and $\Valuations$, $\emptyset$ are the bottom and top elements, respectively.
    Thus, the lattice $(\BeliefSets, \subseteq, \cap, \expn, Cn(\emptyset) , \Lang )$ of belief sets and the lattice $(\ClosedSets, \supseteq, \expn_w, \cap_w, \Valuations, \emptyset )$ of closed sets are bounded and complete by definition.
    Moreover, they are isomorphic, due to the bijection shown in Theorem~\ref{propo:sets-bijections}, the antitone relation between $\Pset(\Lang)$ and $\Pset(\Valuations)$ established by Theorem~\ref{thm:galois-connection}, and how they interact with the operators given $X \subseteq \BeliefSets$:
    \[
    \PLtoPV[\bigplus X] = \PLtoPV[ \bigcup X] = \bigcap \PLtoPV[ X ]
    \qquad
    \PLtoPV[\bigcap X] = \PLtoPV[ \PVtoPL( \bigcup \PLtoPV[X]) ] = \bigplus \PLtoPV[X]
    \]

    Consider now $\mathbb{F}$ the first order language over the signature of bounded lattices, together with a binary operator $\circ$. By extending $\BeliefSets$ with a belief set operator $\generico$ and $\ClosedSets$ with a closed world operator $\generico_w$ restricted to $\ClosedSets$, we obtain $\mathbb{F}$-structures.
    Thus, both $\generico$ and $\generico_w$ admit axiomatic characterizations in $\mathbb{F}$ as interpretations of $\circ$.
    Furthermore, since functions $\PLtoPV$ and $\PVtoPL$ are order-reversing lattice isomorphisms, they can be extended to preserve the logical structure of $\mathbb{F}$, yielding a translation between these two $\mathbb{F}$-structures:
    \[
    \begin{tabular}{ccc}
    $\begin{gathered}
    \PLtoPV[K \subseteq M] = \PLtoPV[M] \subseteq \PLtoPV[K] \\
    \PLtoPV[K \cap M] = Cl(\PLtoPV[K] \cup_w \PLtoPV[M]) \\
    \PLtoPV[K \expn M] = \PLtoPV[K] \cap_w \PLtoPV[M] \\
    \PLtoPV[K \generico M] = \PLtoPV[K] \generico_w \PLtoPV[M]
    \end{gathered}$
    &
    $\begin{gathered}
    \PVtoPL(A \subseteq B) = \PVtoPL(B) \subseteq \PVtoPL(A) \\
    \PVtoPL(A \cap B) = \PVtoPL(A) \expn \PVtoPL(B) \\
    \PVtoPL(Cl(A \cup B)) = \PVtoPL(A) \cap \PVtoPL(B) \\
    \PVtoPL(A \generico_w B) = \PVtoPL(A) \generico \PVtoPL(B)
    \end{gathered}$
    \end{tabular}
    \]
    
    This means that every axiomatic characterization formulated in first order logic over the signature of bounded lattices can be translated between belief set and world operators, without loss of expressive power, via these isomorphisms.
\end{rem}

This remark summarizes how the AWS framework and world versatile operators developed in Section \ref{sec:abstract-world-semantic} acts over a Tarskian logic: every world postulate using $\subseteq$, $\cap_w$, and $\expn_w$, can be seen as elements of $\mathbb{F}$ and translated into belief set postulates using $\subseteq$, $\expn$, and $\cap$ for any Tarskian logic.

But before continuing with the analysis of the axiomatic definition of operators, there is still work to be done.
Belief set operators are not the standard in belief change theory.
Furthermore, this restriction may be originally considered as a particular case of multiple change operators.
However, belief set operators are, in fact, the logical abstraction of belief change operators that satisfy the closure and extensionality postulates of the three classical frameworks.

\begin{propo}\label{propo:extensional-mc-op-are-belief-sets-op}
    Let $\Logic = \langle \Lang, Cn \rangle$ be a Tarskian logic.
    Consider a multiple change operator $\generico: \BeliefSets \times \Pset(\Lang) \rightarrow \Pset(\Lang)$.
    Then, there is a belief set operator $\circ$ where $K \generico M = K \circ Cn(M)$ for every $K \in \BeliefSets$, $M \subseteq \Lang$ iff $\generico$ satisfies:
    \begin{postulate}{MC$\generico$}
    \item \label{post:multiple-change-closure} $Cn(K \generico M) = K \generico M$ (closure).
    \item \label{post:multiple-change-extensionality} If $Cn(M) = Cn(M')$ then $K \generico M = K \generico M'$ (extensionality).
    \end{postulate}
    Moreover, there is a one-to-one correspondence between multiple change operators satisfying \ref{post:multiple-change-closure}$\sim$\ref{post:multiple-change-extensionality} and multiple change operators restricted by $\BeliefSets$.
\end{propo}

\beginpraax{Proposition}{propo:extensional-mc-op-are-belief-sets-op}
We have $K \generico M = K \generico Cn(M)$ by the idempotence of $Cn$ and \ref{post:multiple-change-extensionality}.
Meanwhile, by \ref{post:multiple-change-closure}, the image of $\generico$ belongs to $\BeliefSets$.
Hence, operator $\circ$ is $\generico$ restricted to $\BeliefSets$.
Suppose now $K \generico M = K \circ Cn(M)$ for every $K \in \BeliefSets$, $M \subseteq \Lang$.
Then, $\generico$ satisfies \ref{post:multiple-change-closure}, since the image of $\circ$ belongs to $\BeliefSets$, and \ref{post:multiple-change-extensionality} holds, since $K \generico M = K \circ Cn(M) = K \circ Cn(M') = K \generico M'$ when $Cn(M) = Cn(M')$.

The relation $K \generico M = K \circ Cn(M)$ defines a function from multiple change operators restricted by $\BeliefSets$ to multiple change operators satisfying \ref{post:multiple-change-closure} and \ref{post:multiple-change-extensionality}.
The first part of the previous paragraph shows that this function is surjective.
To see the injectivity, if $\circ$ and $\circ'$ are two multiple change operators restricted by $\BeliefSets$ such that $K \circ Cn(M) = K \generico M = K \circ' Cn(M)$, then $\circ$ and $\circ'$ are the same operator.
Thus, this function is in fact a bijection.
\endpraax

Tarskian logics where $\ClosedSets = \ClopenSets$ ensure us that $\BeliefSets = \BeliefSets_f$ by Proposition \ref{propo:sets-bijections}.
These are Tarskian logics that preserve all the algebraic structure we have in AWS, since every belief set has a complemented counterpart: given $K \in \BeliefSets$, then $\PLtoPV[K]^c \in \ClosedSets$, hence $\neg K = \PVtoPL(\PLtoPV[K]^c) \in \BeliefSets$.
The typical CPL logics satisfying this are the finite ones (i.e., when $\Valuations$ is finite), thus $\Pset(\Valuations) = \ClopenSets$.
Since $\BeliefSets$ is in correspondence to $\Lang$ by the Stone Representation of Boolean Algebras~\cite{Stone1936}, then given $K \in \BeliefSets$ there is a formula $\psi$ such that $K = Cn(\psi)$, and $\neg K = Cn( \neg \psi)$.
So we interpret $\Pset(\Valuations) \times \Pset(\Valuations)$ as $\Lang \times \Lang$, $\cap$ as $\wedge$, the $\cup$ as $\vee$ and the complement as negation. This means that every world operator, since it is naturally closed, can be translated into the KM framework.

\begin{propo}\label{propo:finite-CPL-only-has-world-based-operators}
    Let $\Logic = \langle \Lang, Cn \rangle$ be a finite CPL logic and $\generico: \Lang \times \Lang \longrightarrow \Lang$ be a belief change operator.
    Then, $(\Pset(\Valuations), \cap, \cup, ^c)$ is isomorphic to $(\Logic, \wedge , \vee , \neg)$ and 
    there is a belief set operator $\circ$ such that $Cn(\psi \generico \mu) = Cn(\psi) \circ Cn(\mu)$ for every $\psi$, $\mu \in \Lang$ iff $\generico$ satisfies:
    \begin{postulate}{KM$\generico$}
    \item \label{post:KM-extensionality} If $\psi \equiv \phi$, $\mu \equiv \nu$ then $\psi \generico \mu \equiv \phi \generico \nu$ (extensionality).
    \end{postulate}
    Moreover, this relation defines a one-to-one correspondence.
\end{propo}

\beginpraax{Proposition}{propo:finite-CPL-only-has-world-based-operators}
If $\Logic$ is a CPL logic, by the Stone Representation of Boolean Algebras, we know that $\ClopenSets$ is in correspondence with $\Logic$, which can be interpreted as the set $\mathcal{F} = \{ K \in \BeliefSets \tq $there is $\psi \in \Lang$ such that $K=Cn(\psi) \}$.
Note also that:
\[
\|\neg \psi\| = \| \psi \|^c \quad , \quad \| \psi \wedge \phi \| = \|\psi\| \cup \|\phi\| \quad , \quad \| \psi \vee \phi \| = \|\psi\| \cap \|\phi\|
\]

Since $\Logic$ is T1 Tarskian logic, every finite set of $\Valuations$ is closed.
But $\Valuations$ is finite, thus $\Pset(\Valuations) = \ClosedSets = \ClopenSets$.
Therefore, $\BeliefSets = \mathcal{F} = \BeliefSets_f$ by Proposition \ref{propo:sets-bijections}.
We conclude then that $(\Pset(\Valuations), \cap, \cup, ^c)$ is isomorphic to $(\Logic, \wedge , \vee , \neg)$.

Assume now $\generico: \Lang \times \Lang \longrightarrow \Lang$ satisfies \ref{post:KM-extensionality}
and define $\circ: \BeliefSets \times \BeliefSets \longrightarrow \BeliefSets$ such that $Cn(\psi) \circ Cn(\mu) = Cn(\psi \generico \mu)$.
Then $\circ$ is well-defined due to \ref{post:KM-extensionality} and the isomorphism between $\BeliefSets$ and $\Logic$.
Consider now $\generico: \Lang \times \Lang \longrightarrow \Lang$ where there is a belief set operator 
such that $Cn(\psi \generico \mu) = Cn(\psi) \circ Cn(\mu)$.
Then $Cn(\psi \generico \mu) = Cn(\psi) \circ Cn(\mu) = Cn(\phi) \circ Cn(\nu) = Cn(\phi \generico \nu)$
when $Cn(\psi) = Cn(\phi)$ and $Cn(\mu) = Cn(\nu)$, i.e. $\psi \equiv \phi$ and $\mu \equiv \nu$.
Thus $\generico$ satisfies \ref{post:KM-extensionality}.
\endpraax

In CPL, belief set operators restricted in their second parameter to $\BeliefSets_f$ are equivalent to AGM operators, since the Stone correspondence is also valid when $\Valuations$ is infinite.
This can be directly deduced from Propositions \ref{propo:extensional-mc-op-are-belief-sets-op} and \ref{propo:finite-CPL-only-has-world-based-operators}.

\begin{cor}\label{cor:extensional-AGM-op-are-belief-sets-op}
    Let $\Logic = \langle \Lang, Cn \rangle$ be a CPL logic and $\generico: \BeliefSets \times \Lang \longrightarrow \Pset(\Lang)$ be a belief change operator.
    Then, there is a belief set operator $\circ$ where its second parameter is restricted to $\BeliefSets_f$ such that $K \generico \mu = K \circ Cn(\mu)$ for every $K \in \BeliefSets$ and $\mu \in \Lang$ iff $\generico$ satisfies:
    \begin{postulate}{AGM$\generico$}
    \item \label{post:AGM-revision-closure} $Cn(K \generico \mu) = K \generico \mu$ (closure).
    \item \label{post:AGM-revision-extensionality} If $\mu \equiv \nu$ then $K \generico \mu = K \generico \nu$ (extensionality).
    \end{postulate}
    Moreover, this relation defines a one-to-one correspondence.
\end{cor}

These results show that closure and extensionality postulates, either in the versions of AGM, KM, or Multiple Change frameworks, characterize the belief change operators that connect with the world operators of the AWS framework, i.e., belief set operators.
Through Remark \ref{rem:galois-connection-extended-to-postulates}, this means that we can translate world postulates to belief change postulates when closure and extensionality postulates are also assumed.
Note that the world postulates we are referring to are the one that uses $\subseteq$ and $\cap_w$ since $\expn_w$ is never considered in our proposal, i.e., all of them except for \ref{post:world-shared-success}.
Also, translating postulates is not enough to compare world-selection functions constructions and classical constructions.
This forces us to analyze Tarskian logics according to their representations, focusing on how the translation can capture the union of $\Pset(\ClosedSets)$, the elements of $\Valuations$, and AGM-compliance.

\subsection{A classification of Tarskian logics}

The interpretation of Proposition~\ref{propo:sets-bijections} as lattice isomorphisms allows us to refine the construction of Tarskian representations.
If the elements of the lattice $\BeliefSets$ are intersections of elements of $\MaxConsSets$, then it is enough to consider the set of characteristic functions of $\MaxConsSets$ as $\Valuations$ for a Tarskian representation in Suszko’s reduction.
In particular, every singleton is closed.
A similar result holds when the elements of the lattice are intersections of $\cap$-prime theories.

\begin{propo}\label{propo:topological-prime-tarskian-logics-char}
    Let $\Logic = \langle \Lang, Cn \rangle$ be a Tarskian logic and $\LogRepr = \langle \Lang, \Valuations \rangle$ its Tarskian representation. The following are equivalent:
    \begin{enumerate}
        \item \label{item:topological-closure} $Cl(A \cup B ) = Cl(A) \cup Cl(B)$ for every $A$, $B \subseteq \Valuations$;
        \item \label{item:intersection-union-relation} $\PLtoPV[M \cap N] = \PLtoPV[M] \cup \PLtoPV[N]$ for every $M$, $N \in \BeliefSets$;
        \item \label{item:primeSets-Valuations-relation} If $v \in \Valuations$ then $T_v$ is a prime theory of $\BeliefSets$, i.e., it satisfies:
        \[
        \si M \cap N \subseteq T_v \tonce M\subseteq T_v \mbox{ or } N \subseteq T_v \qquad \forall \, M, \, N \in \BeliefSets
        \]
    \end{enumerate}
    In this case, $(\BeliefSets, \cap, +, Cn(\emptyset), \Lang )$ is distributive\footnote{A fourth equivalence can be given affirming that the lattice of $\BeliefSets$ is a spatial frame~\cite{PassmoreTesis2025}, which implies that it is distributive. However, adding this result was beyond the scope of this work.}, i.e., given $K$, $M$, $N \in \BeliefSets$:
    \[
    N \cap (K + M) = (N \cap K) + (N \cap M) \qquad N + (K \cap M) = (N + K) \cap (N + M)
    \]
\end{propo}

\beginpraax{Proposition}{propo:topological-prime-tarskian-logics-char}
    \begin{description}
        \item[(\ref{item:topological-closure}) iff (\ref{item:intersection-union-relation})] If $M$, $N \in \BeliefSets$, by Theorem \ref{thm:galois-connection}, we have:
        \[
        Cl(\PLtoPV[M] \cup \PLtoPV[N]) = \PLtoPV[ \PVtoPL(\PLtoPV[M] \cup \PLtoPV[N]) ] = \PLtoPV[ \PVtoPL(\PLtoPV[M]) \cap \PVtoPL(\PLtoPV[N]) ] = \|M \cap N\|
        \]
        Hence, if $Cl(A \cup B) = Cl(A) \cup Cl(B)$ for every $A$, $B \subseteq \Valuations$, then:
        \[
        \PLtoPV[M \cap N] = Cl(\PLtoPV[M] \cup \PLtoPV[N]) = Cl(\PLtoPV[M]) \cup Cl(\PLtoPV[N]) = \PLtoPV[M] \cup \PLtoPV[N]
        \]
        Now, if $A$, $B \subseteq \Valuations$, by applying the properties of Theorem \ref{thm:galois-connection}, we have:
        \[
        Cl(A \cup B) = \PLtoPV[\PVtoPL(A \cup B)] = \PLtoPV[\PVtoPL(A) \cap \PVtoPL(B) ] 
        \]
        If $\|M \cap N\| = \PLtoPV[M] \cup \|N\|$ for every $M$, $N \in \BeliefSets$, and $\PVtoPL(A)$ and $\PVtoPL(B) \in \BeliefSets$:
        \[
        Cl(A \cup B) = \PLtoPV[ \PVtoPL(A) \cap \PVtoPL(B) ] = \PLtoPV[ \PVtoPL(A) ] \cup \PLtoPV[ \PVtoPL(B) ] = Cl(A) \cup Cl(B)
        \]
        \item[(\ref{item:intersection-union-relation}) iff (\ref{item:primeSets-Valuations-relation})] We check that $T_v$ is a prime theory for a given $v \in \Valuations$.
        Take $M$, $N \in \BeliefSets$ such that $M \cap N \subseteq T_v$.
        By the Galois Connection of Theorem $\ref{thm:galois-connection}$, $v \in \PLtoPV[M \cap N]$.
        Then, $v \in \PLtoPV[M] \cup \PLtoPV[N]$, meaning $M \subseteq T_v$ or $N \subseteq T_v$.

        Assume now that every $T_v$ is a prime theory, and take $M$, $N \in \BeliefSets$.
        If $v \in \PLtoPV[M \cap N]$, it means that $M \cap N \subseteq T_v$.
        By hypothesis, we have that $M \subseteq T_v$ or $N \subseteq T_v$, hence $v \in \PLtoPV[M] \cup \PLtoPV[N]$.
        We conclude that $\PLtoPV[M \cap N] = \PLtoPV[M] \cup \PLtoPV[N]$.
        The other inclusion is given by Theorem \ref{thm:galois-connection}.
    \end{description}
    Lastly, we show that the lattice $(\BeliefSets, \cap, +, Cn(\emptyset) , \Lang )$ is distributive by applying the properties of Theorem \ref{thm:galois-connection} and item \ref{item:intersection-union-relation}:
    \[
    \begin{array}{ll}
        N \cap (K + M)  & = Cn(N) \cap Cn(K \cup M) = \PVtoPL( \PLtoPV[N] \cup \PLtoPV[K \cup M] ) \\
                        & = \PVtoPL( \PLtoPV[N \cap (K \cup M)] ) = \PVtoPL( \PLtoPV[(N \cap K) \cup (N \cap M)] ) \\
                        & = Cn( (N \cap K) \cup (N \cap M) ) = (N \cap K) + (N \cap M) \\
        N + (K \cap M)  & = Cn( N \cup (K \cap M) ) = Cn( (N \cup K) \cap (N \cup M) ) \\
                        & = \PVtoPL( \PLtoPV[ N \cup K ] \cup \PLtoPV[ N \cup M ] ) = Cn( N \cup K ) \cap Cn( N \cup M ) \\
                        & = ( N + K ) \cap ( N + M )
    \end{array}
    \]
\endpraax

Proposition \ref{propo:topological-prime-tarskian-logics-char} characterizes the logics that have a representation with a topological closure. 
This result allows us to reinterpret Remark \ref{rem:galois-connection-extended-to-postulates} w.r.t a refined version of isomorphisms since $ \PLtoPV[ K \cap M ] = \PLtoPV[ K ] \cup \PLtoPV[ M ]$, thus $\cap$ in $\BeliefSets$ is based on $\cup$ of $\ClosedSets$.
This means that $Cl$ is unnecessary for the translation, as long as the set of belief sets involved is finite.
In particular, in this kind of logics it is possible to translate both the construction using world-selection functions and postulate \ref{post:world-shared-success}, and therefore define versatile operators axiomatically.

The notion of decomposable logic introduced by Flouris et al.~\cite{FPA2006} can also be reinterpreted in terms of the representation of a logic, as it is fundamentally formulated at the level of belief sets.
Decomposable logics characterize those Tarskian logics that admit contraction operators, that is, AGM-compliant logics.

\begin{defn}[\cite{FPA2006}]\label{def:flouris-complement-beliefs}
    Let $\Logic = \langle \Lang, Cn \rangle$ be a Tarskian logic.
    Given $K \in \BeliefSets$ and $M \subseteq \Lang$, we define $M^-(K)$ the set of complement beliefs of $M$ w.r.t. $K$ as:
    \[
    M^-(K) \! = \!\! \begin{cases}
        \{X \subseteq \Lang \tq Cn(X) \!\subsetneq K , \, Cn(M\cup X)=K \} & \si Cn(\emptyset) \subsetneq Cn(M) \subseteq K \\
        \{X \subseteq \Lang \tq Cn(X)=K \} & \cc
    \end{cases}
    \]
    The set $M^-(K)$ can be extended to any $K \subseteq \Lang$ as $M^-(K) = M^-(Cn(K))$.
    We say $\Logic$ is decomposable if for every $K$, $M \subseteq \Lang$ we have that $M^-(K) \neq \emptyset$.
\end{defn}

Let us classify Tarskian representations (and therefore their logics) in terms of properties of the closure operator $Cl$, according to these properties.

\begin{defn}\label{defn:Tarskian-classification}
    Let $\LogRepr = \langle \Lang, \Valuations \rangle$ be a Tarskian representation. We say $\LogRepr$ is:
\begin{description}
    \item[\textbf{complete}] \label{complete-condition} if every singleton of $\Valuations$ is a closed set;
    \item[\textbf{topological}] \label{topological-condition} if $Cl(A\cup B)= Cl(A) \cup Cl(B)$ for every $A$, $B \subseteq \Valuations$;%\footnote{the name comes by the fact that $Cl$ is a topological closure};
    \item[\textbf{decomposable}] \label{decomposable-condition} if given $B \in \ClosedSets$ and $A \subseteq B$ such that $B \neq \Valuations$, there is $v \not \in B$ such that $B \cap Cl(A \cup \{v\}) =Cl(A)$;
    \item[\textbf{detachable}] \label{detachable-condition} if given $B \in \ClosedSets$, $B \neq \Valuations$, there is $v \not \in B$ such that $ B \cap Cl(\{v\}) = \emptyset$.
\end{description}
\end{defn}

The next theorem presents how these characterizations relates with each other.

\begin{thm}\label{thm:valuations-set-characterization}
    Let $\Logic = \langle \Lang, Cn \rangle$ be a Tarskian logic and $\LogRepr = \langle \Lang, \Valuations \rangle$ its Tarskian representation.
    Then:
\begin{enumerate}
    \item If $\LogRepr$ is complete, then it is detachable.
    \item $\Logic$ is decomposable iff $\LogRepr$ is decomposable. In particular, $\LogRepr$ is detachable.
    \item If $\LogRepr$ is topological and detachable then $\Logic$ is decomposable.
\end{enumerate}
\end{thm}

\beginpraax{Theorem}{thm:valuations-set-characterization}
\begin{enumerate}
    \item If $\LogRepr$ is complete, then $\{v\} = Cl(\{v\})$. Hence, for every $B \neq \Valuations$ there is $v \not \in B$ such that $B \cap Cl(\{v\}) = B \cap \{v\} = \emptyset$, i.e., $\LogRepr$ is detachable.

    \item Let $K \in \BeliefSets$ and $X \subseteq K$ such that $Cn(\emptyset) \neq Cn(X)$.
    If $Cn(X) = Cn(K)$, take $Z=\emptyset$ and then $Cn(X \cup Z)=Cn(K)$.
    So assume $Cn(X) \neq Cn(K)$ and take $A=\PLtoPV[K]$ and $B = \| X \| = \| Cn(X) \|$.
    Since $\PLtoPV$ is antitone, we have $A=\PLtoPV[K] \subsetneq \|X \| = B \neq \Valuations$.
    If $\LogRepr$ is decomposable, there is some $v \not \in B$ such that $B \cap Cl(A \cup \{v\}) = Cl(A) = A$.
    Consider then $Z=\PVtoPL(A \cup \{v\})$.
    Note that $Cn(Z)=Z \subsetneq K$, since $A=\PLtoPV[K] \subsetneq \|Z\|=Cl(A \cup \{v\})$ and $v \not \in A$.
    Also, $\|X \cup Z\| = \|X\| \cap \|Z\| = B \cap Cl(A \cup \{v\}) = A$, therefore $Cn(X\cup Z) = K$ when applying $\PVtoPL$, i.e., $\Logic$ is decomposable.

    Consider $B \in \ClosedSets$, $B \neq \Valuations$, and $A \subseteq B$, and take $K=\PVtoPL(A)$ and $X=\PVtoPL(B)$.
    Since $\PVtoPL$ is antitone, we have $X \subseteq K$.
    If $\Logic$ is decomposable, there is $Z \subseteq \Lang$ where $Cn(Z) \subsetneq K$ and $Cn(X \cup Z) = Cn(K)=K$.
    If $C=\|Z\| \in \ClosedSets$, then $Cl(A) \subsetneq C$ and $A =\PLtoPV[K] = \|X \cup Z\| = \|X\| \cap \|Z\| = B \cap C$.
    Take $v \in C$, $v \not \in Cl(A)$.
    Since $C \in \ClosedSets$, then $Cl(A) \subsetneq Cl(A \cup \{v\}) \subseteq C$.
    Hence, $Cl(A) \subseteq B \cap Cl(A \cup \{v\}) \subseteq B \cap C = A$.
    We deduce $Cl(A) = B \cap Cl(A \cup \{v\})$, and $v \not \in B$ since $v \not \in Cl(A)$.
    We conclude $\LogRepr$ is decomposable.
    Also, if $\LogRepr$ is decomposable, then it is detachable by considering $A=\emptyset$ for every $B \neq \Valuations$.
    
    \item Take $A$, $B \in \ClosedSets$ such that $A \subseteq B \neq \Valuations$.
    If $\LogRepr$ is detachable, there is $v \not \in B$ where $Cl(\{v\}) \cap B = \emptyset$.
    If $\LogRepr$ is topological, $Cl(A \cup \{v\})= A \cup Cl(\{v\}) \in \ClosedSets$, therefore $B \cap (A \cup Cl(\{v\})) = A$, i.e., $\LogRepr$ is decomposable.
\end{enumerate}
\endpraax

Some known logics already satisfy many of these properties.
For example, if it is finitary, then there is a complete Tarskian representation due to Lindenbaum's lemma~\cite{Lindstrom2022}.
In the case of CPL, there is a complete and topological representation: the one associated with the Stone Space~\cite{Stone1936}.
This is also related to having classical disjunction and negation in a Tarskian logic, as it is seen in Lindström~\cite{Lindstrom2022}, Ribeiro et al.~\cite{RNW18nonCompact} and Sauerwald \cite{Sauerwald2024}.

\begin{rem}\label{obs:topological-complete-logic-example}
    Let $\LogRepr = \langle \Lang, \Valuations \rangle$ be a Tarskian representation.
    \begin{enumerate}
        \item If $\Logic$ has a 2-arity symbol $\vee$ such that $\| \alpha \vee \beta \| = \|\alpha \| \cup \|\beta\|$ for every $\alpha$, $\beta \in \Lang$, i.e., $v(\alpha \vee \beta)=\max\{ v(\alpha) , v(\beta) \}$ for every $v \in \Valuations$, then $\LogRepr$ is topological.
        \item If $\Logic$ has a 1-arity symbol $\neg$ such that $\| \neg \alpha \| = \|\alpha \|^c$ for every $\alpha \in \Lang$, i.e., every $v \in \Valuations$ satisfies that $v(\neg \alpha)=1-v(\alpha) $, then $\LogRepr$ is complete.
    \end{enumerate}
\end{rem}

\beginpraax{Remark}{obs:topological-complete-logic-example}
\begin{enumerate}
    \item If $v \in Cl(A) \cup Cl(B)$ then $v \in Cl(A \cup B)$ by the inclusion property of $Cl$.
Take now $v \not \in Cl(A) \cup Cl(B)$, then $\PVtoPL(A) \not \subseteq T_v$ and $\PVtoPL(B) \not \subseteq T_v$.
Hence, there is $\alpha \in \PVtoPL(A)$ and $\beta \in \PVtoPL(B)$ such that $\alpha$, $\beta \not \in T_v$, i.e., $v(\alpha) = v(\beta) = 0$.
By hypothesis, $v(\alpha \vee \beta) = 0$, thus $v \not \in \PLtoPV[\{\alpha \vee \beta\}]$.
Meanwhile, we have $A \subseteq \| \alpha \|$ and $B \subseteq \| \beta \|$ by the Galois Connection.
Then, by hypothesis $A \cup B \subseteq \| \alpha \vee \beta \|$. 
Since $\| \alpha \vee \beta \| \in \ClosedSets$, we have $Cl(A \cup B) \subseteq \| \alpha \vee \beta \|$.
Combining it with the fact that $v \not \in \| \alpha \vee \beta \|$, we conclude $v \not \in Cl(A \cup B)$.
Therefore $Cl(A \cup B) = Cl(A) \cup Cl(B)$.

\item Given $v \in \Valuations$, we show that $\PVtoPL(\{v\})$ is a maximally consistent set.
Take an arbitrary $\alpha \in \Lang$, then either $v \in \PLtoPV[\{\alpha\})]$ or $v \in \PLtoPV[\{\alpha\})]^c$.
Therefore, $Cl(\{v\}) \subseteq \PLtoPV[\{\alpha\})]$ or $Cl(\{v\}) \subseteq \PLtoPV[\{\alpha\})]^c$, or equivalently,
$Cl(\{v\}) \cap \PLtoPV[\{\alpha\})]$ is either $Cl(\{v\})$ or $\emptyset$.
Since, $Cl(\{v\}) \cap \PLtoPV[\{\alpha\})] = \PLtoPV[ \PVtoPL(\{v\}) \cup \{\alpha\} ]$, by applying $\PVtoPL$ we have that $Cn(\PVtoPL(\{v\}) \cup \{\alpha\})$ is either $\PVtoPL(\{v\})$ or $\Lang$.
Therefore, $\PVtoPL(\{v\}) \in \MaxConsSets$, and by the bijection of Proposition \ref{propo:sets-bijections}, $Cl(\{v\}) = \{v\}$.
\end{enumerate}
\endpraax

We end this subsection by defining the logics usually considered in our work.

\begin{defn}\label{def:T1-tarskian-logics}
    We say $\Logic = \langle \Lang, Cn \rangle$ is a T1 Tarskian logic if it has a Tarskian representation $\LogRepr = \langle \Lang, \Valuations \rangle$ that is simultaneously topological and complete, i.e., $\langle \Valuations, Cl \rangle$ is a T1 topological space~\cite{munkres2017topology}.
    Moreover, whenever a Tarskian logic $\Logic$ is referred to by a name introduced in Definition~\ref{defn:Tarskian-classification}, it is understood that $\Logic$ admits a Tarskian representation $\LogRepr$ with this property.
\end{defn}

By Theorem \ref{thm:valuations-set-characterization}, we deduce that T1 Tarskian logics are AGM-compliant logics.
Moreover, when the construction on Remark \ref{rem:galois-connection-extended-to-postulates} is considered in these logics, further correspondences follow naturally.
Given that the representation is topological, the operations $\cup_w$ and $\expn_w$ on $\ClosedSets$ coincide when applied to finite families of sets, i.e., $\cup_w$ in $\ClosedSets$ is finitely associated with $\cap$ in $\BeliefSets$. Thus, belief set operators may be constructed using world-selection functions.
Furthermore, the set of valuations $\Valuations$ is associated with $\MaxConsSets$ since the representation is complete, providing a more fine-grained interpretation of world-selection.

This class includes CPL logics, any finitary logic with a classical disjunction, and any logic with both a classical negation and a classical disjunction.
A detailed analysis of which logics satisfy the T1 condition lies beyond the scope of this work; related investigations can be found, for instance, in Ribeiro’s thesis~\cite{Ribeiro23} and in Guimarães et al.~\cite{GuimaOzakiRibeiro23}.
Our primary objective, however, remains the use of AWS to represent the three main frameworks of belief change theory within CPL.
For this reason, results are stated in terms of complete, topological, or T1 Tarskian logics, preserving the level of generality achieved.
In the next section, we exploit the systematic translation to define Multiple Change Versatile Operators, and show how world-selection constructions capture classical multiple change constructions. 

\section{Multiple Change Versatile Operators}\label{sec:mc-versatile-op}

Section~\ref{sec:AWS-applied-TLogics}, via Remark~\ref{rem:galois-connection-extended-to-postulates}, establishes a systematic translation of the AWS framework into belief set operators.
Since $\ClosedSets$ need not be closed under union and complement, these operations were deliberately avoided in most world postulates formulated in Section~\ref{sec:abstract-world-semantic}.
Those world postulates were stated to be translated into belief set postulates, enabling the definition of belief set operators even in non-AGM-compliant scenarios.
What follows is the collection of belief change postulates obtained by applying this translation.
\begin{itemize}
    \item Postulates to define contraction operators:
\begin{postulate}{$\cont$}
    \item $K \cont M \subseteq K$ (contraction inclusion). \label{post:c-inclusion}
    \item $K \subseteq (K \cont M) \expn M$ (recovery). \label{post:c-recovery}
    \item \label{post:c-success} If $M \neq Cn(\emptyset)$ then $M \not \subseteq K \cont M$ (contraction success).
    \item \label{post:c-vacuity} If $M \not \subseteq K$ then $K \cont M = K$ (contraction vacuity).
\end{postulate}
    \item Postulates to define revision operators:
\begin{postulate}{$\rev$}
    \item $M \subseteq K \rev M$ (revision success). \label{post:r-success}
    \item $K \rev M \subseteq K \expn M$ (inclusion). \label{post:r-inclusion}
    \item \label{post:r-consistency} If $M \neq \Lang$ then $K \rev M \neq \Lang$ (revision consistency).
    \item \label{post:r-vacuity} If $K \expn M \neq \Lang$ then $K \rev M = K \expn M$ (revision vacuity).
\end{postulate}
    \item Postulates to define versatile operators:
\begin{postulate}{$\vers$}
    \item $K \cap M \subseteq K \vers M$ (shared success). \label{post:shared-success}
    \setcounter{postcounter}{2}
    \item \label{post:credible-consistency} If $M \neq \Lang$ then $(K \vers M) \expn M \neq \Lang$ (credible consistency).
    \item \label{post:normal-credible-vacuity} If $K \expn M \neq \Lang$ then $K \subseteq K \vers M$ (normal credible vacuity).
\end{postulate}
    \item Postulates associated with relations:
\begin{postulate}{Rel}
    \item \label{post:credible-vacuity} If $K \expn M \neq \Lang$ then $(K \generico M) \expn M \neq \Lang$ (credible vacuity).
    \item \label{post:normal-questionable-vacuity} If $K \expn M \neq \Lang$ then $M \subseteq (K \generico M) \expn K$ (questionable vacuity).
    \item \label{post:full-knowledge-preservation} If $M \not \subseteq (K \vers M) \expn K $ then $K \vers M \subseteq K$ (knowledge preservation).
    \item \label{post:consistency-preservation} If $K \neq \Lang$ then $K \vers M \neq \Lang$ (consistency preservation).
    \item \label{post:weak-relative-success} Either $M \subseteq K \vers M$ or $K \subseteq K \vers M$ (weak relative success).
\end{postulate}
\end{itemize}

In any case, it is possible to define belief set operators on weaker conditions than T1 Tarskian logics.
However, as expected, we may need stronger assumptions to construct such operators using world-selection functions.

\begin{thm}\label{thm:closed-world-op-compliant-repr}
    Let $\Logic$ be a topological Tarskian logic, and $\cont_w$,  $\rev_w$, $\vers_w$ be elementary world contraction, revision, and versatile operators, respectively. Then:
    \begin{enumerate}
        \item $\cont_w$ is closed iff $\sigma(A,B) = Cl(\sigma(A,B)) \cap A^c$ for every $A$, $B \in \ClosedSets$.
        \item $\rev_w$ is closed iff $\sigma(A,B^c) = Cl(\sigma(A,B^c)) \cap A^c$ for every $A$, $B \in \ClosedSets$.
        \item $\vers_w$ is closed iff its associated revision operators are closed.
    \end{enumerate}
    In this context, we say a world-selection function $\sigma$ is $\generico_w$\textbf{-compliant} if $\generico_w$ is a closed elementary world contraction or revision constructed by $\sigma$.
\end{thm}

\beginpraax{Theorem}{thm:closed-world-op-compliant-repr}

Having $\cont_w$ closed means that $A \cup \sigma(A,B) \in \ClosedSets$ for every $A$, $B \in \ClosedSets$.
Since $\Logic$ is topological $A \cup \sigma(A,B) = Cl(A \cup \sigma(A,B)) = A \cup Cl(\sigma(A,B))$.
Thus, $\sigma(A,B) = Cl(\sigma(A,B)) \cap A^c$.
Assume now $\sigma(A,B) = Cl(\sigma(A,B)) \cap A^c$.
Thus, we have $A \cup \sigma(A,B) = A \cup Cl(\sigma(A,B))$.
Since $\Logic$ is topological, if $A \in \ClosedSets$ then $Cl(A \cup \sigma(A,B)) = A \cup Cl(\sigma(A,B)) = A \cup \sigma(A,B)$.
Therefore, $\cont_w$ is closed.
    
Analogously, if $\rev_w$ closed, $(A \cap B) \cup \sigma(A,B^c) \in \ClosedSets$ for every $A$, $B \in \ClosedSets$.
Then $(A \cap B) \cup \sigma(A,B^c) = (A \cap B) \cup Cl(\sigma(A,B^c))$ since $\Logic$ is topological and $A \cap B \in \ClosedSets$.
Thus, $\sigma(A,B) = Cl(\sigma(A,B)) \cap A^c$.
Assume $\sigma(A,B^c) = Cl(\sigma(A,B^c)) \cap A^c$.
Then, $A \cup \sigma(A,B^c) = A \cup Cl(\sigma(A,B^c))$.
Note $Cl(\sigma(A,B^c)) \subseteq B$ if $B \in \ClosedSets$.
Hence $(A\cap B) \cup \sigma(A,B^c) = (A\cap B) \cup Cl(\sigma(A,B^c))$.
Since $\Logic$ is topological, $Cl((A\cap B) \cup \sigma(A,B^c)) = (A\cap B) \cup \sigma(A,B^c)$ when $A \in \ClosedSets$.
Thus, $\rev_w$ is closed.

Recall $A \vers_w B = (A \rev_w^\sigma B) \cup (B \rev_w^\rho A)$ by the Theorem \ref{thm:world-versatile-separation-representation}.
If $\vers_w$ is closed, given $A$, $B \in \ClosedSets$ we know that $A \vers_w B \in \ClosedSets$.
Hence, $A \rev_w^\sigma B = (A \vers_w B) \cap B \in \ClosedSets$ and $B \rev_w^\rho A = (A \vers_w B) \cap A \in \ClosedSets$.
Therefore, $\rev_w^\sigma$ and $\rev_w^\sigma$ are closed operators.
If $\rev_w^\sigma$ and $\rev_w^\sigma$ are closed operators, then $\vers_w$ is a closed operator since $\Logic$ is topological.
\endpraax

Note that properties $\cont_w$-compliant and $\rev_w$-compliant are independent: they only overlap when $B \in \ClopenSets$.
Therefore, some world-selection functions can satisfy both.
Previous result can be reinterpreted to characterize the values of compliant world-selection functions when working with a topological Tarskian Logic.

\begin{cor}\label{cor:top-val-are-almost-compliant}
    Let $\Logic$ be a topological Tarskian logic, $\sigma$ be a world-selection function and $\sigma(A,B) \in \Sigma_{(A,B)} = \{S \subseteq B^c \cap A^c \tq A \cup S \in \ClosedSets \}$.
    If $A$, $B \in \ClosedSets$, then:
    \begin{enumerate}
        \item $\sigma$ is $\cont_w$-compliant iff $\sigma(A,B) \in \Sigma_{(A,B)}$;
        \item $\sigma$ is $\rev_w$-compliant iff $\sigma(A,B^c) \in \Sigma_{(A,B^c)}$.
    \end{enumerate}
    Also, if $\Logic$ is T1, every finite subset of $B^c \cap A^c$ is in $\Sigma_{(A,B)}$.
\end{cor}

\beginpraax{Corollary}{cor:top-val-are-almost-compliant}
The items are directly deduced by Theorem \ref{thm:closed-world-op-compliant-repr}.
By working with T1 Tarskian logics, we can ensure that every finite subset of $\Valuations$ is closed.
This allows us to affirm that every finite subset of $B^c \cap A^c$ is a possible value of $\sigma(A, B)$.
\endpraax

This result is key to perceiving the diversity of belief set operators in topological Tarskian logics.
Moreover, T1 Tarskian logics are not only AGM-compliant, but capable of generating the diverse family of belief change operators usually seen in CPL, now through world-selection functions.

\subsection{Operators beyond AGM-compliance}
Since postulate \ref{post:shared-success} is implied by either \ref{post:c-inclusion}, \ref{post:r-success}, or \ref{post:weak-relative-success}, this suggests operators based on elementary world contraction, revision, and CL revision can be defined in any Tarskian logic.

\begin{thm}\label{thm:logic-elem-rev-cont-CL-repr-theorem}
    Given $\cont$, $\rev$ and $\vers$ belief set operators in a Tarskian logic $\Logic$:
    \begin{enumerate}
        \item $\cont$ satisfies \ref{post:c-inclusion} and \ref{post:c-recovery} iff it is based on an elementary world contraction $\cont_w$.
        Moreover, in this case, if $\Logic$ is topological and $\cont_w$ is constructed by $\sigma$, then $K \cont M = K \cap \PVtoPL(\sigma(\PLtoPV[K], \PLtoPV[M]))$.

        We call $\cont$ an elementary (belief sets) contraction, constructed by $\sigma$.
        \item $\rev$ satisfies \ref{post:r-success} and \ref{post:r-inclusion} iff it is based on an elementary world revision $\rev_w$.
        Moreover, in this case, if $\Logic$ is topological and $\rev_w$ is constructed by $\sigma$, then
        $K \rev M = (K + M) \cap \PVtoPL(\sigma(\PLtoPV[K], \PLtoPV[M]^c))$.
        
        We call $\rev$ an elementary (belief sets) revision, constructed by $\sigma$.
        \item $\vers$ satisfies \ref{post:shared-success}, \ref{post:r-inclusion}, \ref{post:full-knowledge-preservation} and \ref{post:weak-relative-success} iff it is based on an elementary world CL revision $\vers_w$, iff it is constructed as an \textit{elementary (belief sets) CL revision}, i.e., by a relation $\rel$ and an elementary revision $\rev$ as follows:
        \[
        K \vers M = \begin{cases}
            K \rev M & \si K \rel M \\
            K & \cc
        \end{cases}
        \]
    \end{enumerate}
\end{thm}

\beginpraax{Theorem}{thm:logic-elem-rev-cont-CL-repr-theorem}
    The equivalences are given by the results that provide the extension of the isomorphisms of Definition \ref{rem:galois-connection-extended-to-postulates}. 
    In the context of topological Tarskian logic, the construction of the world operators, defined using the world-selection and $\cup$ of $\ClosedSets$, allows us to see the belief set operators as constructed by the world-selection and $\cap$ of $\BeliefSets$.
    The relation from $\ClosedSets$ can also be naturally extended, giving the elementary CL revision structure.
\endpraax

We avoid writing the proofs of the following representation theorems of this subsection since they are all analogous to this one.

When it comes to contraction operators, this theorem puts the recovery postulate in the non-AGM-compliant logics scenario~\cite{flouris2006}.
This is valid since the incompatibility occurs when both contraction success and recovery postulates hold.
It also shows that, in topological Tarskian logics, elementary operators have axiomatic and constructive presentations, although possibly trivial.

Note that an operator constructed by intersecting two elementary revisions satisfies elementary versatile postulates in every Tarskian logic.
However, no representation theorem links this construction with the axiomatic definition.
This result holds when $\BeliefSets$ is a distributive lattice, as in Theorem \ref{propo:topological-prime-tarskian-logics-char}.

\begin{thm}\label{thm:logic-elem-vers-repr-theorem}
Let $\vers$ be a belief set operator in a topological Tarskian logic $\Logic$.
Then, the following are equivalent:
\begin{enumerate}
    \item $\vers$ satisfies \ref{post:shared-success} and \ref{post:r-inclusion};
    \item $\vers$ is based on an elementary world versatile $\vers_w$;
    \item $\vers$ is constructed by a pair of elementary revision $\rev^\sigma$ and $\rev^\rho$ as follows:
    \[
    K \vers M = (K \rev^\sigma M) \cap (M \rev^\rho K)
    \]
    \item $\vers$ is constructed by a pair of world-selection functions $(\sigma , \rho)$ as follows:
    \[
    K \vers M = (K + M) \cap \PVtoPL(\sigma(\PLtoPV[K], \PLtoPV[M]^c)) \cap \PVtoPL(\rho(\PLtoPV[M], \PLtoPV[K]^c))
    \]
\end{enumerate}
    We call $\vers$ an elementary (belief sets) versatile, constructed by $(\sigma, \rho)$.
\end{thm}

With elementary versatile operators in topological Tarskian logics, it is also possible to represent other world-based operators, as long as the postulates stay compatible, such as the elementary filtered revision case.

\begin{thm}\label{thm:logic-elem-filtered-revision-repr}
Let $\vers$ be a belief set operator in a topological Tarskian logic $\Logic$.
Then, the following are equivalent:
\begin{enumerate}
    \item $\vers$ satisfies \ref{post:shared-success}, \ref{post:r-inclusion} and \ref{post:full-knowledge-preservation};
    \item $\vers$ is based on an elementary world filtered revision $\vers_w$;
    \item $\vers$ is constructed as an elementary (belief set) filtered revision, i.e., by an elementary revision $\rev$ and two disjoint relations $\rel_{cred}$ and $\rel_{allow}$ as follows:
        \[
        K \vers M = \begin{cases}
            K \rev M & \si K \rel_{cred} M \\
            (K \rev M) \cap K & \si K \rel_{allow} M \\
            K & \cc
        \end{cases}
        \]
\end{enumerate}
\end{thm}

Only the credible and questionable relations remain to be translated from AWS.
To simplify, let us consider the relations associated with elementary versatile operators instead of an arbitrary belief sets one, based on Corollary \ref{cor:vers-cred-quest-relations}.

\begin{defn}\label{def:change-perspective-relations}
    Let $\vers$ be an elementary versatile in a topological Tarskian logic.
    We define the following relations over $\BeliefSets$:
    \begin{itemize}
        \item The relation $K \Cred M$ given by $(K \generico M) + M \neq \Lang$ or $K = \Lang$;
        \item The relation $K \Quest M$ given by $M \not \subseteq (K \generico M) + K $;
        \item The relation $K \FullQuest M$ given by $M \not \subseteq (K \generico M) + K \subseteq K$.
    \end{itemize}
\end{defn}

Hence, if $\vers$ is based on $\vers_w$, then $K \Cred M$ can be defined as $\PLtoPV[K] \Cred_w \PLtoPV[M]$, $K \Quest M$ as $\PLtoPV[K] \Quest_w \PLtoPV[M]$, and $K \FullQuest M$ as $\PLtoPV[K] \FullQuest_w \PLtoPV[M]$.
Also, a belief set version of Proposition \ref{propo:world-connections-cred-quest-relations} can be given, connecting relations with postulates \ref{post:normal-questionable-vacuity}$\sim$\ref{post:weak-relative-success}.
As in AWS, these relations standardize the construction of filtered, CL revision, and shielded contraction.
However, assuming that the associated contraction or revision satisfies their postulates, whether the relation condition is satisfied or not generates a problem when considering basic contraction or revision in a non-AGM-compliant logic.
We amend this by assuming that the associated operators satisfy their postulates constrained by their relations.

\begin{thm}\label{thm:operators-with-relations}
Let $\vers$ be a belief set operator over a topological Tarskian logic.
Then $\vers$ satisfies \ref{post:shared-success}, \ref{post:r-inclusion} and \ref{post:full-knowledge-preservation} iff given $\Cred$ and $\Quest$ the associated relations of $\vers$, there is an elementary revision $\rev$ such that:
    \[
    K \vers M = \begin{cases}
        K \rev M & \si K \not \Quest M\\
        (K \rev M) \cap K & \si K \Cred M \mbox{ and } K \Quest M\\
        K & \si K \not \Cred M \mbox{ and } K \Quest M\\
    \end{cases}
    \]
\end{thm}

The same result is also valid for CL revision.

\begin{thm}\label{thm:logic-cl-revision-repr-theorem}
Let $\vers$ be a belief set operator over a Tarskian logic.
Then $\vers$ satisfies \ref{post:shared-success}, \ref{post:r-inclusion}, \ref{post:full-knowledge-preservation} and \ref{post:weak-relative-success} iff given $\Cred$ and $\Quest$ the associated relations of $\vers$, there is an elementary revision $\rev$ such that:
    \[
    K \vers M = \begin{cases}
        K \rev M & \si K \not \Quest M \\
        K & \si K \Quest M\\
    \end{cases}
    \]
Moreover, in this case:
\begin{enumerate}
    \item $\vers$ also satisfies \ref{post:r-vacuity} and \ref{post:consistency-preservation} iff $\rev$ satisfies \ref{post:r-consistency} and \ref{post:r-vacuity} for all the pairs $(K, M)$ satisfying $K \Cred M$ and $\Cred$ is equivalent to $\not \Quest$.
\end{enumerate}
\end{thm}

Therefore, we covered from elementary to normal CL revision.

Due to the use of the complement in the AWS framework, the construction for shielded contraction needs an extra relation, based on $\Cred$.

\begin{thm}\label{thm:shielded-contraction-repr}
Let $\cont$ be a belief set operator over a topological Tarskian logic. The following are equivalent:
\begin{enumerate}
    \item $\cont$ satisfies \ref{post:c-inclusion} and \ref{post:c-recovery};
    \item there is an elementary world versatile operator $\vers_w$ satisfying \ref{post:world-c-inclusion}, such that $\PLtoPV[ K \cont M ] = \PLtoPV[K] \vers_w \PLtoPV[M]^c$ and its relation $\Cred_w$ satisfies:
    \[
    \PLtoPV[K] \Cred \PLtoPV[M]^c \quad \mbox{ iff } \quad M \not \subseteq K \cont M \mbox{ or } K = \Lang
    \]
    \item $\cont$ is constructed by an elementary contraction $\cont'$ and a relation $\rel$ over $\BeliefSets$ as follows:
    \[
    K \cont M = \begin{cases}
        K \cont' M & \si K \rel M\\
        K & \cc
    \end{cases}
    \]
    where $\cont'$ also satisfies \ref{post:c-success} for all the pairs $(K, M)$ satisfying $K \rel M$.
\end{enumerate}
\end{thm}

This shows non-prioritized operators can be naturally extended to multiple change scenarios in non-AGM-compliant logics.
Prioritized operators, however, differ due to AGM-compliance.
In the next subsection, we consider T1 Tarskian logics, our rich AGM-compliant logics, to define the familiar AGM construction of remainders for choice contraction and package revision.
Moreover, we show this constructions are associated with basic world contraction and revision operators.

\subsection{Classical Constructions in T1 Tarskian Logics}
Remainder-like sets are the foundation of the typical construction in the AGM theory.
Here, we compare choice remainders and package openings of the Multiple Change framework~\cite{Fuhrmann1988, H1993, FH1994} with world-selection functions.
We start by presenting a technical lemma that will help us connect these constructions.

\begin{lem}\label{the-max-max-lemma}
    Let $(X,\leq)$ be a partial order set and let $Y, Z \in \Pset(X)$.
    Consider $Max: \Pset(X) \rightarrow \Pset(X)$ where $Max(R) = \{ r \in R \tq $if $s \in R$ then $r \not < s \}$.

    If $Z \subseteq Y$ and $\emptyset \neq \setMax(Y) \subseteq Z$ then $\emptyset \neq \setMax(Z)$ and $\setMax(Y) = \setMax(Z)$.
\end{lem}

\beginpraax{Lemma}{the-max-max-lemma}
Take $A \in \setMax(Y)$.
Since $Z \subseteq Y$, then every $B \in Z$ satisfies either $B \leq A$ or $B$ is not related to $A$.
Nevertheless, since $\setMax(Y)$ is a subset of $Z$, $A \in Z$.
Hence, $A \in \setMax(Z)$.
For the other inclusion, take $B \in \setMax(Z)$ and assume $B < C$ for some $C \in Y$.
By definition, $A \not < C$ for every $A \in \setMax(Y)$.
Hence, either there is some $A \in \setMax(Y)$ such that $C \leq A$ or every $A \in \setMax(Y)$ does not relate to $C$.
The first case implies that $B < A$ by transitivity, and the latter implies that $C \in \setMax(Y)$, thus $C \leq B$.
Since any case leads to a contradiction, we conclude $B \not < C$ for every $C \in Y$, i.e. $B \in \setMax(Y)$.
\endpraax %% END OF PROOF

The definition of choice remainders for multiple change is an adaptation of the AGM classical remainders.
Here, we add another adjustment to this definition by considering it a two-parameter function.

\begin{defn}[\cite{FH1994}]\label{choice-remainders}
We define the function $\remainder_c : \Pset(\Lang) \times \Pset(\Lang) \longrightarrow \Pset(\Pset(\Lang))$ as choice remainder if for every $X \in K \remainder_c M$:
\begin{postulate}{C$_\bot$}
    \item $X \subseteq K$. \label{choice-rem-inclusion}
    \item $M \not \subseteq Cn(X)$\footnote{Originally $\models$ is extended for sets~\cite{FH1994}. Here we use $\models$ classically, i.e., only formulas are allowed on the right side, and use $Cn$ to deal with sets.}. \label{choice-rem-condition}
    \item If $X \subsetneq X' \subseteq K$ then $M \subseteq Cn(X')$. \label{choice-rem-maximality}
\end{postulate}
\end{defn}

Levi Identity is used to define revision from contraction, by means of a negation operator.
Since the negation of a set is not uniquely defined, some authors defined a new construction that avoids the issue: package openings~\cite{F1996}.
As before, we adapted the opening set definition as a two-parameter function.

\begin{defn}[\cite{F1996}]\label{package-openings}
We define the function $\opening_p : \Pset(\Lang) \times \Pset(\Lang) \longrightarrow \Pset(\Pset(\Lang))$ as package opening if for every $X \in K \opening_p N$:
\begin{postulate}{P$_\opening$}
    \item $X \subseteq K$. \label{package-open-inclusion}
    \item $X \cup N \not \models \bot$. \label{package-open-condition}
    \item If $X \subsetneq X' \subseteq K$ then $X' \cup N \models \bot$. \label{package-open-maximality}
\end{postulate}
\end{defn}

Note that $N \cup X \models \bot$ iff $\PLtoPV[N] \cap \PLtoPV[X] = \emptyset$ iff $\PLtoPV[X] \subseteq \PLtoPV[N]^c$.
Analogously, $M \subseteq Cn(X)$ iff $\PLtoPV[X] \subseteq \PLtoPV[M]$.
Hence, if $M, N \subseteq \Pset(\Lang)$ are such that $\PLtoPV[N]^c = \PLtoPV[M]$ (in particular, $Cn(M)$ and $Cn(N) \in \BeliefSets_f$), then $K \remainder_c M = K \opening_p N$.
The idea outlined above suggests the following theorem.

\begin{thm}\label{thm:remainders-openings-characterization}
Let $\Logic$ be a T1 Tarskian logic. Then, given $K \in \BeliefSets$, $M \subseteq \Lang$:
\[
K \remainder_c M = \setMax(\lbrace K \cap T_v \, |  v \not \in \PLtoPV[M] \rbrace) \qquad K \opening_p M = \setMax(\lbrace K \cap T_v \, |  v \in \PLtoPV[M] \rbrace)
\]
Moreover, $K \remainder_c M$ and $K \opening_p M$ are subsets of $\BeliefSets$.
\end{thm}

\beginpraax{Theorem}{thm:remainders-openings-characterization}
Take $S = \lbrace X \, | X \subseteq K \,, \, M \not \subseteq Cn(X) \rbrace$ and $S' = \lbrace K  \cap T_v \tq v \not \in \PLtoPV[M]  \rbrace$ (resp. $S=\lbrace X \, | X \subseteq K \,, \, X \cup M \not \models \bot \rbrace$ and $S' = \lbrace K \cap T_v \tq v \in \PLtoPV[M] \rbrace$) to cover both proofs simultaneously, since they are essentially identical.

Let us start working with $S'$.
Note $S' \subseteq \BeliefSets$ since $K$, $T_v \in \BeliefSets$, implies $K \cap T_v \in \BeliefSets$.
Also note $M \subseteq K$ (resp. $K \cup M \models \bot$) implies $\setMax(S') = S'$.
Since the logic is T1 Tarskian, the elements in $S'$ are incomparable over inclusion:
If $K \cap T_v \subseteq K \cap T_w \neq K$, then $\PLtoPV[K] \neq \PLtoPV[K \cap T_w] \subseteq \PLtoPV[K \cap T_v]$ by the Galois Connection.
Hence $\PLtoPV[K] \neq \PLtoPV[K] \cup \{w\} \subseteq \PLtoPV[K] \cup \{v\}$, since the logic is topological and complete.
We conclude that $v$, $w \not \in K$ and thus $v=w$.
Therefore $\setMax(S') = S'$.

Then, we prove $\setMax(S) = \setMax(S')$.
We first show $S' \subseteq S$.
This is trivial when $\PLtoPV[M]^c = \emptyset$ (resp. $\PLtoPV[M] = \emptyset$).
Otherwise, take $v \not \in \PLtoPV[M]$ (resp. $v \in \PLtoPV[M]$) and consider $X = K \cap T_v$.
Then $X \subseteq K$ and $Cn(X) \subseteq T_v$.
Since $M \not \subseteq T_v$ (resp. $M \subseteq T_v$), we have that $M \not \subseteq Cn(X)$ (resp. $M \cup X \not \models \bot$).
Therefore, $X \in S$.

To use Lemma \ref{the-max-max-lemma}, we also need to show $\emptyset \neq \setMax(S) \subseteq S'$.
If $\emptyset \neq \setMax(S)$, consider $X \in S$ and take $v \in \PLtoPV[X] \cap \PLtoPV[M]^c$ (resp. $\PLtoPV[X] \cap \PLtoPV[M]$).
Note there is always such $v$, otherwise $X$ would fail \ref{choice-rem-condition} (resp. \ref{package-open-condition}).
Then, $X \subseteq K \cap T_v$ by \ref{choice-rem-inclusion} (resp. \ref{package-open-inclusion}).
Suppose $X \subsetneq K \cap T_v$, then by \ref{choice-rem-maximality} (resp. \ref{package-open-maximality}) we deduce $M \subseteq Cn(K \cap T_v)$ (resp. $M \cup (K \cap T_v) \models \bot$). But $Cn(K \cap T_v) \subseteq T_v$ and $M \not \subseteq T_v$ (resp. $M \subseteq T_v$), leading to a contradiction.
Thus $X = K \cap T_v \in S'$.

Lastly, if $\setMax(S) = \emptyset$, then $K \not \in S$ by \ref{choice-rem-maximality} (resp. \ref{package-open-maximality}).
Thus $M \subseteq K$ (resp. $K \cup M \models \bot$), so we affirm $\setMax(S') = S'$.
Assume $M \not \subseteq Cn(\emptyset)$ (resp. $M \not \models \bot$), and take $v \in \PLtoPV[M]^c$ (resp. $v \in \PLtoPV[M]$). Hence $K \cap T_v \in S'$.
Since $S' \subseteq S$ and $S' = \setMax(S')$, then $\setMax(S) \neq \emptyset$, leading to a contradiction.
Thus $M \subseteq Cn(\emptyset)$ (resp. $M \models \bot$).
Therefore $S' = \emptyset$.
In any case, $\setMax(S) = \setMax(S')$.
\endpraax % END OF PROOF

Theorem \ref{thm:remainders-openings-characterization} confirms that both constructions behave like the AGM remainder set $K \remainder \alpha$ when $K$ is a belief set.
Specifically, it generalizes the results where remainder sets are belief sets~\cite{TextbookHansson1999}, and there is a bijection between the remainder sets and the set of possible worlds that do not entail the new information~\cite{Grove1988}.
Also, Theorem \ref{thm:remainders-openings-characterization} implies these constructions satisfy extensionality.

\begin{cor}\label{cor:remainder-opening-extensional-properties}
    Let $K \in \BeliefSets$, $M$, $M' \subseteq \Lang$:
    \begin{enumerate}
        \item $K \remainder_c M = K \remainder_c Cn(M)$ and $K \opening_p M = K \opening_p Cn(M)$.
        \item \label{cc-pw-M-extensionality} If $M \subseteq K$ and $K \remainder_c M = K \remainder_c M'$ then $Cn(M)=Cn(M')$.
        \item \label{po-pw-M-extensionality} If $K \cup M \models \bot$ and $K \opening_p M = K \opening_p M'$ then $Cn(M)=Cn(M')$.
        \item \label{cc-pw-remainder-empty-case} $M \subseteq Cn(\emptyset)$ iff $K \remainder_c M = \emptyset$.
        \item \label{po-pw-remainder-empty-case} $M \models \bot$ iff $K \opening_p M = \emptyset$.
    \end{enumerate}
\end{cor}

Continuing with the construction of classical operators, we introduce Hansson's two-place selection function~\cite{H1993} to preserve the two-parameter representation.

\begin{defn}[\cite{H1993}]\label{def:selection-function}
Define $\gamma: \Pset(\Lang) \times \Pset(\Pset(\Lang)) \rightarrow \Pset(\Pset(\Lang))$ the two-place selection function such that
$\gamma(N,S) \subseteq S$, $\gamma(N,\emptyset)=N$ and $\gamma(N,S)\neq \emptyset$ if $S \neq \emptyset$.
\end{defn}

Through Theorem \ref{thm:remainders-openings-characterization} and Corollary \ref{cor:remainder-opening-extensional-properties}, we can relate selection functions over choice remainders or package openings to world-selection functions.

\begin{defn}\label{def:classical-world-selection-function}
    Let $\Logic$ be a Tarskian logic.
    We say a world-selection function $\sigma$ is \textit{classical} if it is $\cont$-compliant, $\rev$-compliant, normal, and semantically successful.
    A classical world-selection function $\sigma$ and a selection function $\gamma$ are paired if for every $K \in \BeliefSets$, $M \subseteq \Lang$ they satisfy:
    \begin{itemize}
        \item $\PLtoPV[\bigcap \gamma(K, K \remainder_c M) ] = \PLtoPV[K] \cup \sigma(\PLtoPV[K], \PLtoPV[M])$;
        \item $\PLtoPV[\bigcap \gamma(K, K \opening_p M) ] = \PLtoPV[K] \cup \sigma(\PLtoPV[K], \PLtoPV[M]^c)$.
    \end{itemize}
\end{defn}

In T1 Tarskian logics, this relation suggests that selection functions and classical world-selection represent the same behavior, i.e., they are interchangeable.

\begin{propo}\label{propo:classic-selection-as-world-selection}
    If $\Logic$ is a T1 Tarskian logic, then every selection function has a paired classical world-selection function, and every classical world-selection function has a paired selection function.
\end{propo}

\beginpraax{Proposition}{propo:classic-selection-as-world-selection}
    Assume $\gamma$ is a selection function and define $\sigma$ as follows:
    \[
    \sigma(A,B) = \begin{cases}
        \PLtoPV[\bigcap \gamma(\PVtoPL(A), \PVtoPL(A) \remainder_c \PVtoPL(B)) ] \cap A^c & \si A, B \in \ClosedSets\\
        \PLtoPV[\bigcap \gamma(\PVtoPL(A), \PVtoPL(A) \opening_p \PVtoPL(B^c)) ] \cap A^c & \si A, B^c \in \ClosedSets\\
        \sigma'(A,B) & \cc
    \end{cases}
    \]
    where $\sigma'(A,B)$ is a normal and semantically successful world-selection function.
    Such $\sigma'$ always exists since it is not restricted to the constraints of $\gamma$ or $\Lang$.
    Also, if $A \in \ClosedSets$ and $B \in \ClopenSets$, then by Theorem \ref{thm:remainders-openings-characterization}:
    \[
    \PVtoPL(A) \remainder_c \PVtoPL(B) = \setMax(\lbrace \PVtoPL(A) \cap T_v \, |  v \in B^c \rbrace) = \PVtoPL(A) \opening_p \PVtoPL(B^c)
    \]
    Therefore, $\sigma$ is well-defined.
    In particular, if $B = \Interpretations$, then $B \in \ClopenSets$.
    Since $\setMax(\lbrace \PVtoPL(A) \cap T_v \, |  v \in B^c \rbrace) = \emptyset$, we deduce $\sigma(A,B) = \PLtoPV[\Lang] \cap A^c = \emptyset$.

    From now on, assume $A$, $B \in \ClosedSets$ and $B \neq \Interpretations$ to show all the properties of $\sigma$.
    If $A \subseteq B$, then $\PVtoPL(A) \remainder_c \PVtoPL(B) = \{ \PVtoPL(A \cup \{v\}) \tq v \not \in B \} \neq \emptyset$.
    By definition of $\gamma$, $\Logic$ being complete and the property of $\PVtoPL$ for arbitrary intersections, there is some $\emptyset \neq S \subseteq B^c$ such that $\PLtoPV[\bigcap \gamma(\PVtoPL(A), \PVtoPL(A) \remainder_c \PVtoPL(B)) ] = Cl(A \cup S)$.
    Therefore, $\sigma(A,B) = Cl(A \cup S) \cap A^c$ and $\sigma(A,B) \cup A= Cl(A \cup S)$.
    Note that $A \subseteq B \subseteq S^c$, hence $S \subseteq A^c$.
    So we can conclude $\emptyset \neq \sigma(A,B) \subseteq A^c \cap B^c$ and $\sigma(A,B) \cup A \in \ClosedSets$.
    
    Analogously, if $A \cap B = \emptyset$, then $\PVtoPL(A) \opening_p \PVtoPL(B)) = \{ \PVtoPL(A \cup \{v\}) \tq v \in B \} \neq \emptyset$.
    Hence, there is $\emptyset \neq S \subseteq B$ such that $\PLtoPV[\bigcap \gamma(\PVtoPL(A), \PVtoPL(A) \opening_p \PVtoPL(B)) ] = Cl(A \cup S)$.
    Therefore, $\sigma(A,B^c) = Cl(A \cup S) \cap A^c$ and $\sigma(A,B^c) \cup A= Cl(A \cup S)$.
    Then $S \subseteq A^c$ since $A \subseteq B^c \subseteq S^c$.
    We conclude $\emptyset \neq \sigma(A,B^c) \subseteq A^c \cap B$ and $\sigma(A,B^c) \cup A \in \ClosedSets$.
    
    Assume now $A \not \subseteq B$, then $\PVtoPL(A) \remainder_c \PVtoPL(B)) = \{ \PVtoPL(A) \}$.
    Hence, by definition of $\gamma$, $\sigma(A,B) = Cl(A) \cap A^c$.
    Since $A \in \ClosedSets$, we conclude $\sigma(A,B) = \emptyset$.
    Analogously, if $A \cap B \neq \emptyset$, then $\PVtoPL(A) \opening_p \PVtoPL(B)) = \{ \PVtoPL(A) \}$.
    As before, we conclude $\sigma(A,B^c) = \emptyset$.

    Therefore, we covered the normal and semantically successful cases, and simultaneously showed that $\sigma$ is a world-selection function, and that $\sigma(A, B) \cup A$, $\sigma(A, B^c) \cup A \in \ClosedSets$, i.e., $\sigma$ is $\cont_w$-compliant and $\rev_w$-compliant by Corollary \ref{cor:top-val-are-almost-compliant}.
    Since these cases cover all the possible scenarios, we showed that $\sigma$ is classical.
    Moreover, we showed $\sigma$ is paired with $\gamma$ when we deduced $\sigma(A, B) \cup A= Cl(A \cup S)$ in the remainder case, and $\sigma(A, B^c) \cup A= Cl(A \cup S)$ in the opening case.

    Assume now $\sigma$ is a classical world-selection function and, given $\gamma'$ an arbitrary selection function, define $\gamma$ as follows:
    \[
    \gamma(K,X) = \begin{cases}
        \{K \cap T_v \tq v \in \sigma(\PLtoPV[K], \PLtoPV[M]) \} & \si K \in \BeliefSets \, , \, X \neq \{K\} \, ,\\
        & \exists M \tq X = K \remainder_c M\\
        \{ K \cap T_v \tq v \in \sigma(\PLtoPV[K], \PLtoPV[N]^c) \} & \si K \in \BeliefSets \, , \, X \neq \{K\} \, ,\\
        & \exists N \tq X = K \opening_p N\\
        \gamma'(K,X) & \cc
    \end{cases}
    \]
    Such $\gamma'$ always exists since it is not restricted to the constraints of $\sigma$.
    In particular, $\gamma(K,\{K\}) = \gamma'(K,\{K\}) = \{K\}$ by definition.
    To confirm the well-definition of $\gamma$, first consider the case $K \in \BeliefSets$ and $X = \emptyset$.
    By taking $M=Cn(\emptyset)$ and $N=\Lang$ we have $X = K \remainder_c M = K \opening_p N$, and $\Valuations = \PLtoPV[M] = \PLtoPV[N]^c$.
    Since $\sigma$ is a world-selection function, we have $\sigma(\PLtoPV[K], \Valuations) = \emptyset$.
    Therefore, $\gamma(K,\emptyset) = \emptyset$ in both cases.
    If $X \neq \emptyset$, $K \not \in X $ and there are $M$, $N \subseteq \Lang$ such that $X = K \remainder_c M = K \opening_p N$, then by Theorem \ref{thm:remainders-openings-characterization} we deduce $\PLtoPV[K] \subseteq \PLtoPV[M] = \PLtoPV[N]^c$.
    Thus, the sets are the same, and $\gamma$ is well-defined.
    
    Now we show that $\gamma$ is a selection function: take $K \in \BeliefSets$ and $X$ such that $X \neq \emptyset$ and $X \neq \{K\}$ and there is $M \subseteq \Lang$ where either $X= K \remainder_c M$ or $X= K \opening_p M$.
    This means either $M \subseteq K$ or $K \cup M \models \bot$, i.e. $\PLtoPV[K] \subseteq \PLtoPV[M]$ or $\PLtoPV[K] \subseteq \PLtoPV[M]^c$.
    By theorem \ref{thm:remainders-openings-characterization} and the fact that $\sigma$ is semantically successful, we deduce that either $\emptyset \neq \gamma(K,X) \subseteq K \remainder_c M$, since $\emptyset \neq \sigma(\PLtoPV[K], \PLtoPV[M]) \subseteq \PLtoPV[M]^c$, or $\emptyset \neq \gamma(K,X) \subseteq K \opening N$ since $\emptyset \neq \sigma(\PLtoPV[K], \PLtoPV[M]^c) \subseteq \PLtoPV[M]$.
    In any case, we conclude $\gamma$ is a selection function.

    Lastly, we show $\gamma$ is paired with $\sigma$. If $K \in \BeliefSets$, $M \subseteq \Lang$ and $K \remainder_c M \neq \{K\}$, then, by applying $\gamma$ definition, $\Logic$ is topological, and $\sigma$ is $\cont$-compliant:
    \begin{multline*}
        \PLtoPV[\bigcap \gamma(K, K \remainder_c M) ] = \PLtoPV[K \cap \PVtoPL(\sigma(\PLtoPV[K], \PLtoPV[M])) ] = \PLtoPV[K] \cup Cl(\sigma(\PLtoPV[K], \PLtoPV[M])) \\= \PLtoPV[K] \cup \sigma(\PLtoPV[K], \PLtoPV[M])
    \end{multline*}
    Analogously, if $K \in \BeliefSets$, $N \subseteq \Lang$, and $K \opening_p N \neq \{K\}$, then, by applying $\gamma$ definition, $\Logic$ is topological, and $\sigma$ is $\rev$-compliant:
    \begin{multline*}
        \PLtoPV[\bigcap \gamma(K, K \opening_p N) ] = \PLtoPV[K \cap \PVtoPL(\sigma(\PLtoPV[K], \PLtoPV[N]^c)) ] = \PLtoPV[K] \cup Cl(\sigma(\PLtoPV[K], \PLtoPV[N]^c)) \\
        = \PLtoPV[K] \cup \sigma(\PLtoPV[K], \PLtoPV[N]^c)
    \end{multline*}
    The case where either $K \remainder_c M = \{K\}$ or $K \opening_p N = \{K\}$, we have that $\PLtoPV[\bigcap \gamma(K, \{K\}) ] = \PLtoPV[K]$.
    These cases imply either $M \not \subseteq K$ or $K \cup N \not \models \bot$.
    Therefore, $\sigma(\PLtoPV[K], \PLtoPV[M]) = \emptyset$ or $\sigma(\PLtoPV[K], \PLtoPV[N]^c) = \emptyset$, by $\sigma$ being normal.
\endpraax

Since choice contraction and package revision also generalize the constructions in AGM and KM, this result shows that classical world-selection functions fully cover the constructions for the three classical frameworks.

\begin{cor}\label{cor:choice-contraction-package-revision-are-world-based}
    Let $\Logic$ be a T1 tarskian logic, we define belief set operators $-_c$ and $\rev_p$ respectively as choice contraction and package revision if:
    \[
    K -_c M = \bigcap \gamma(K,K \remainder_c M) \qquad K \rev_p N = \bigcap \gamma(K,K \opening_p N) + N
    \]

    Then, $\cont$ is a choice contraction iff $\cont$ satisfies \ref{post:c-inclusion}$\sim$\ref{post:c-vacuity}, iff $\cont$ is based on $\cont_w$ a basic world contraction.
    Analogously, $\rev$ is a package revision iff $\rev$ satisfies \ref{post:r-success}$\sim$\ref{post:r-vacuity}, iff $\rev$ is based on $\rev_w$ a basic world revision.
\end{cor}

Lastly, we present the remaining non-prioritized operator, moderated revision.

\begin{cor}\label{thm:multilple-mod-revision-representation}
    Given $\vers$ a belief set operator in a T1 Tarskian logic $\Logic$. The following are equivalent:
    \begin{enumerate}
        \item $\vers$ satisfies \ref{post:shared-success}, \ref{post:credible-consistency} and \ref{post:r-vacuity};
        \item $\vers$ is based on a world moderated revision $\vers_w$;
        \item $\vers$ is constructed by a belief sets package revision $\rev^\sigma$ and a belief sets elementary revision $\rev^\rho$ that satisfies \ref{post:r-vacuity} as $K \vers M = (K \rev^\sigma M) \cap (M \rev^\rho K)$.
    \end{enumerate}
\end{cor}

With this, the versatile operators of the AWS framework were presented in different logic environments, but focusing on the classical frameworks for CPL.

\section{Conclusions And Future Work}\label{sec:conclusions} % 19 lineas para 35 páginas

In this work, we developed the theory of belief change in an Abstract Worlds Semantics (AWS) framework.
This is a purely semantic foundation for belief change, where worlds are treated as primitives, and the algebra of sets of worlds provides the underlying structure.
Within this framework, we reformulated both classical and non-prioritized operators and unified them under a broader family of operators called versatile.
We also standardized the notions of credibility and questionability, originally introduced in non-prioritized operators such as shielded contraction, filtered, CL, and moderated revision, by presenting them as relations instead of sets of formulas.

By connecting AWS to Tarskian logics through logical valuations, we showed that this semantics successfully captures the three classical belief change frameworks —AGM, KM, and Multiple Change— when restricted to belief sets.
We presented a formal, systematic translation by which all the world versatile operators defined in AWS can be equivalently interpreted within any of these three frameworks, both axiomatically and constructively.
Moreover, since this association applies to all Tarskian logics, it allowed us to analyze the possibility of defining such operators in other logics.
Thus, through this translation, we could affirm the belief change theory can be generalized beyond AGM-compliant logics.

This work also opens new perspectives on several long-standing ideas and problems.
The very definition of versatile operators reshapes how the Levi/Harper duality relates to non-prioritized operators, while the translation developed offers a clearer understanding of this duality within the multiple change framework.
Also, by showing that versatile operators, heavily based on the recovery postulate, can be defined in non-AGM-compliant logics, we extend and reinterpret the results of AGM-compliance~\cite{flouris2006}.
Lastly, the standardization of credibility and questionability concepts as relations based on the behavior of any operator allows us to analyze operators in terms of these concepts, beyond the original intention of their definition.

The AWS framework thus establishes a foundation for a research program rather than a single formal result.
In Grimaldi's thesis~\cite{GrimTesis2025}, this program was developed more broadly: it included the analysis of non-prioritized operator subfamilies, the adaptation of supplementary postulates, and the incorporation of additional operators such as update, erase~\cite{KM1991}, local promotion~\cite{GMR2021}, and CL update~\cite{FKPS2023-CLupd}.
We avoided these aspects here to keep our proposal distinct from the extensive existing work on possible world semantics, but they constitute the natural continuation of our proposal.
In the same sense, we deliberately circumvented specific non-classical logics that have been extensively studied in belief change theory, such as modal logic, paraconsistent logics, description logics, or Horn logics.
This decision was made because our primary objective was to demonstrate that our proposal simultaneously encompasses the three CPL classical frameworks, rather than to provide a thorough analysis of non-prioritized operators in non-classical logics.
Nevertheless, we are aware of these developments, and we expect to apply our systematic translation in future works for analyzing such cases.
Lastly, we acknowledge that this proposal does not account for knowledge-base operators, and such an adaptation toward belief base change theory remains to be explored in depth as future work.

\begin{acks}
Martinez was partially supported by the Spanish MCIN/AEI (CHIST-ERA iTrust) project
PCI2022-135010-2, project PID2022-139835NB-C21, and PIE 2023-5AT010 CSIC.
Grimaldi and Rodriguez acknowledge partial support of Argentinean project UBA-CyT-20020190100021BA, and the MOSAIC project from the European Union’s Horizon 2020 research and innovation programme under the Marie Sklodowska-Curie grant agreement No 10100762.
\end{acks}

\printbibliography

\newpage

\appendix

\section{Proofs}

\appendixcontent

\end{document}